\documentclass[preprint,3p,times,twocolumn,sort&compress]{elsarticle}

\pdfoutput=1
\usepackage{lineno}
\modulolinenumbers[5]

\journal{ArXiv}

\usepackage{url}
\usepackage[breaklinks, hidelinks]{hyperref}

\usepackage{graphicx}

\usepackage{algpseudocode}
\usepackage{algorithm}
\makeatletter
\newcommand{\algorithmicmargin}{\the\ALG@thistlm}
\makeatother
\newcommand{\obj}[1]{\textbf{\textit{#1}}}
\newcommand{\method}[3]{\obj{#1}.\Call{#2}{#3}}
\newcommand{\var}[2][{}]{\ensuremath{\mathtt{#2}\ifthenelse{\equal{#1}{}}{}{=#1}}}

\newcommand{\alglines}[3]{Alg.~\ref{#1}, lines~\ref{#2}--\ref{#3}}
\algnewcommand\True{\textbf{true}}
\algnewcommand\False{\textbf{false}}
\algnewcommand{\Break}{\textbf{break}}
\algnewcommand{\Wrap}{\par\hspace{\dimexpr\algorithmicmargin-0.5em}$\hookrightarrow$\enskip}

\usepackage{amsmath}
\usepackage{mathtools}
\newcommand{\ceil}[1]{\left \lceil #1 \right \rceil }

\renewcommand{\vec}[1]{\mathbf{#1}}

\usepackage{multirow}
\usepackage{booktabs}
\usepackage{makecell}

\usepackage[format=hang, labelformat=parens]{subcaption}
\usepackage{siunitx}
\DeclareSIUnit{\sample}{S}
\DeclareSIUnit{\SamplePerSecond}{SPS}

\usepackage{pdflscape}

\usepackage{neuralnetwork}

\usepackage[acronym]{glossaries}
\setacronymstyle{short-long}
\newacronym{HEP}{HEP}{High Energy Physics}
\newacronym{LHC}{LHC}{Large Hadron Collider}
\newacronym{CERN}{CERN}{the European Organization for Nuclear Research}
\newacronym{PM}{PM}{Post Mortem}
\newacronym{PMS}{PM System}{\glsentrylong{PM} System}
\newacronym{PMF}{PM Framework}{\glsentrylong{PM} Framework}
\newacronym{LEP}{LEP}{Large Electron-Positron Collider}
\newacronym{PS}{PS}{Proton Synchrotron}
\newacronym{SPS}{SPS}{Super Proton Synchrotron}
\newacronym{ATLAS}{ATLAS}{A Toroidal LHC ApparatuS}
\newacronym{CMS}{CMS}{Compact Muon Solenoid}
\newacronym{ALICE}{ALICE}{A Large Ion Collider Experiment} 
\newacronym{LHCb}{LHCb}{Large Hadron Collider beauty experiment}
\newacronym{SUSY}{SUSY}{SUperSYmmetry}
\newacronym{SSB}{SSB}{Spontaneous Symmetry Breaking}
\newacronym{QPS}{QPS}{Quench Protection System}
\newacronym{MPS}{MPS}{Machine Protection System}
\newacronym{SDDS}{SDDS}{Self-Describing Data Set}
\newacronym{MTF}{MTF}{Magnet Test Folder}
\newacronym{ELQA}{ELQA}{ELectrical Quality Assurance}
\newacronym{CALS}{CALS}{CERN Accelerators Logging Service}
\newacronym{SM}{SM}{Standard Model}
\newacronym{FCC}{FCC}{Future Circular Collider}
\newacronym{FCC-hh}{FCC-hh}{\glsentryshort{FCC} - proton-proton collider}
\newacronym{FCC-ee}{FCC-ee}{\glsentryshort{FCC} - electron-positron collider}
\newacronym{EE}{EE}{Energy Extraction}
\newacronym{TE-MPE}{TE-MPE}{Technology Department - Machine Protection and Electrical Integrity}
\newacronym{TE-MPE-EE}{TE-MPE-EE}{\glsentrylong{TE-MPE} section - Electrical Engineering}
\newacronym{TE-MPE-EP}{TE-MPE-EP}{\glsentrylong{TE-MPE} section - Electronics for Protection}

\newacronym{HL}{HiLumi}{High Luminosity}
\newacronym{HL-LHC}{HL-LHC}{\glsentrylong{HL} LHC}
\newacronym{ADC}{ADC}{Analogue to Digital Converter}
\newacronym{SAR}{SAR}{Successive Approximation Register}
\newacronym{uQDS}{uQDS}{universal \glsentrylong{QPS}}
\newacronym{DCCT}{DCCT}{Direct-Current Current-Transformer}

\newacronym{DL}{DL}{Deep Learning}
\newacronym{LSTM}{LSTM}{Long Short-Term Memory}
\newacronym{GRU}{GRU}{Gated Recurrent Unit}
\newacronym{RMSE}{RMSE}{Root-Mean-Square Error}
\newacronym{NN}{NN}{Neural Network}
\newacronym{RNN}{RNN}{Recurrent \glsentrylong{NN}}
\newacronym{FNN}{FNN}{Feed-forward \glsentrylong{NN}}
\newacronym{CNN}{CNN}{Convolutional \glsentrylong{NN}}
\newacronym{RL}{RL}{Reinforcement Learning}
\newacronym{HTM}{HTM}{Hierarchical Temporal Memory}

\newacronym{ML}{ML}{Machine Learning}
\newacronym{SVM}{SVM}{Support Vector Machine}
\newacronym{OC-SVM}{OC-SVM}{One Class \glsentrylong{SVM}}
\newacronym{RBF}{RBF}{Radial Basis Function}
\newacronym{ARIMA}{ARIMA}{Autoregressive Integrated Moving Average}
\newacronym{EGADS}{EGADS}{Extendible Generic Anomaly Detection System}
\newacronym{RPCA}{RPCA}{Robust Principle Component Analysis}

\newacronym{ASIC}{ASIC}{Application-Specific Integrated Circuit}
\newacronym{FPGA}{FPGA}{Field-Programmable Gate Array}
\newacronym{HDL}{HDL}{Hardware Description Language}
\makeglossaries

\usepackage{easy-todo}

\usepackage[T1]{fontenc}
\newcommand*\chem[1]{\ensuremath{\mathrm{#1}}}
\begin{document}

\begin{frontmatter}

\title{The model of an anomaly detector for HiLumi LHC magnets based on Recurrent Neural Networks and adaptive quantization}

\author[agh_eit]{Maciej Wielgosz}
\ead{wielgosz@agh.edu.pl}

\author[alma_mater]{Matej Mertik}
\ead{matej.mertik@almamater.si}

\author[agh_fis]{Andrzej Skocze\'n}
\ead{skoczen@fis.agh.edu.pl}

\author[cern]{Ernesto De Matteis}
\ead{ernesto.de.matteis@cern.ch}

\address[agh_eit]{Faculty of Computer Science, Electronics and Telecommunications, AGH University of Science and Technology, Krak\'ow, Poland}
\address[alma_mater]{Alma Mater Europaea - European Center Maribor, Slovenia}
\address[agh_fis]{Faculty of Physics and Applied Computer Science, AGH University of Science and Technology, Krak\'ow, Poland}
\address[cern]{The European Organization for Nuclear Research - CERN, CH-1211 Geneva 23 Switzerland}

\begin{abstract}

This paper focuses on an examination of an applicability of Recurrent Neural Network models for detecting anomalous behavior of the CERN superconducting magnets.
In order to conduct the experiments, the authors designed and implemented an adaptive signal quantization algorithm and a custom GRU-based detector and developed a method for the detector parameters selection.

Three different datasets were used for testing the detector.
Two artificially generated datasets were used to assess the raw performance of the system whereas the 231 MB dataset composed of the signals acquired from HiLumi magnets was intended for real-life experiments and model training.
Several different setups of the developed anomaly detection system were evaluated and compared with state-of-the-art OC-SVM reference model operating on the same data.
The OC-SVM model was equipped with a rich set of feature extractors accounting for a range of the input signal properties.

It was determined in the course of the experiments that the detector, along with its supporting design methodology, reaches F1 equal or very close to 1 for almost all test sets.
Due to the profile of the data, the \var{best\_length} setup of the detector turned out to perform the best among all five tested configuration schemes of the detection system.
The quantization parameters have the biggest impact on the overall performance of the detector with the best values of input/output grid equal to 16 and 8, respectively.
The proposed solution of the detection significantly outperformed OC-SVM-based detector in most of the cases, with much more stable performance across all the datasets.

\end{abstract}


\end{frontmatter}

\section{Introduction}
\label{section:intro}

\begin{table*}
\caption{Comparison of main parameters of \glsentryshort{LHC} \cite{evans2008lhc}, \glsentryshort{HL-LHC} \cite{HL-LHC_pre_drep} and \glsentryshort{FCC-hh} \cite{FCC_params}.} \label{tab:hl-lhc-params}
\centering
\begin{tabular}{lccccc}
\toprule
&Perimeter&Particle energy&Luminosity&Integrated luminosity&\multirow{2}{*}{Number of bunches}\\
&[\si{\kilo\meter}]&[\si{\tera\electronvolt}]&[\SI{e34}{\per\square\centi\meter\per\square\second}]&[\si{\per\femto\barn\per\day}]& \\
\midrule
\glsentryshort{LHC} & \num{27} & \num{7} & \num{1.0} & \num{0.47} & \num{2808} \\
\glsentryshort{HL-LHC} & \num{27} & \num{7} & \num{5.0} & \num{2.8}$^\dagger$ & \num{2748} \\
\glsentryshort{FCC-hh} & \num{100} & \num{50} & \num{5.0} & \num{2.2} & \num{10600} \\
\bottomrule
\end{tabular}
\flushleft
\footnotesize{$^\dagger$ Base design value without taking into account any enhancenet in beam instrumentation.}
\end{table*}

The \gls{LHC} was built with more than 20 years lasted effort of \gls{CERN} personnel and whole worldwide \glsentrylong{HEP} community.
The \gls{LHC} consists of a \SI{27}{\kilo\metre} ring located \SI{100}{\meter} underground and filled mainly with superconducting magnets.
The \gls{LHC} started operating in 2008, and since that time it contributed to some pronounced scientific discoveries concerning \glsentrylong{SM} \cite{atlas2012observation, cms2012observation}.

Many research and development programs are continuously carried out to deliver improvements to the construction of numerous subsystems of the \gls{LHC}.
Currently, the most critical project, which already entered a construction phase, is the \gls{HL-LHC}.
This major upgrade is planned to be introduced between years 2023 and 2025.
The primary goal is to increase the luminosity (rate of collisions) by a factor of five beyond its original design value, and the integrated (over a whole year) luminosity by a factor of ten \cite{HL-LHC_pre_drep}.
It will be possible thanks to the development of several innovative technologies, mainly in the field of superconductivity. 

Looking far into the future, the \gls{CERN} started a study for next generation circular accelerator called \gls{FCC} \citep{FCC_params}.
Preliminary assumptions concerning this project indicate that the \gls{FCC} will be a ring with a circumference around four times longer than the tunnel currently used by \gls{LHC} (Tab.~\ref{tab:hl-lhc-params}).
In such a vast project the importance of an intelligent automation will be much higher and may even be the only chance to maintain and operate the accelerator.

The modifications in the accelerator structure related to the \gls{HL-LHC} project require, in turn, a creation of new solutions for the \gls{MPS}, the \gls{LHC} components maintenance and monitoring system, which is the responsibility of \gls{CERN} \gls{TE-MPE} group. The main interest of the TE-MPE group is to maximise the availability of the machine while the high safety level is guaranteed. 

A complexity of this task stems from an abundance of signals acquired from various \gls{LHC} magnets and the real-time operation requirement.
The system needs to process all the data and detect anomalies in such a time that will allow various automatic fault prevention procedures to run.
Conventional anomaly detection systems, such as presented in section~\ref{section:state-of-the-art}, cannot be used in this particular application due to a vast quantity of signals, very few anomaly cases and the hardware implementation in embedded systems requirement.

The primary goal of this study is a creation and verification of a model that can be used in the future design of a software and ultimately a hardware solution for an anomaly detection device suitable for application in particle accelerators.
One of the promising research directions involves using \gls{NN} and \gls{ML} algorithms for magnets monitoring, as well as anomaly detection.
A real-time execution of \gls{ML} algorithms requires dedicated, low latency architectures, such as \gls{FPGA} or a digital \gls{ASIC}, which is an authors' long-term research goal.
The current work, presented in this paper, focuses on the development and verification of a dedicated solution involving adaptive quantization and \gls{RNN}.
The created solution achieved very encouraging results for \gls{LHC} magnets signals.

The presented research main contributions are as follows:
\begin{itemize}
\item development of an architecture for anomaly detection based on \gls{GRU},
\item introduction of a new approach based on adaptive grid quantization,
\item detector design procedure which accounts for the detector operation environment,
\item development of a system level model suited for doing experiments with the adaptive grid-based approach; the software is available online \citep{online:bitbucket:anomaly_detection}.
\end{itemize}

The developed design procedure should allow reusing the researched solution for various use cases, requiring only a setup configuration changes.

The rest of the paper is organized as follows.
Sections~\ref{section:lhc} and \ref{section:state-of-the-art} provide background information about \gls{LHC} and anomaly detection state of the art (including \glsentrylongpl{RNN} usage), respectively.
The quantization algorithm, the system description, and the developed methodology are explained in sections \ref{section:grid}, \ref{section:system-desc}, and \ref{section:methodology}.
Next, the results of the experiments and their comparison to an alternative approach based on \gls{OC-SVM} method are presented in section~\ref{section:experiments}. General discussion can be found in section~\ref{section:discussion}.
Finally, the conclusions of our research are presented in section~\ref{section:conclusions} and the future work plans in section~\ref{section:future}. Software variables used throughout the whole text are briefly summarized in \ref{appx:soft_symb}.

\section{The \glsentrylong{LHC}}
\label{section:lhc}

\subsection{Superconducting magnets}
\label{section:lhc-magnets}

\begin{figure}
	\includegraphics[width=\columnwidth]{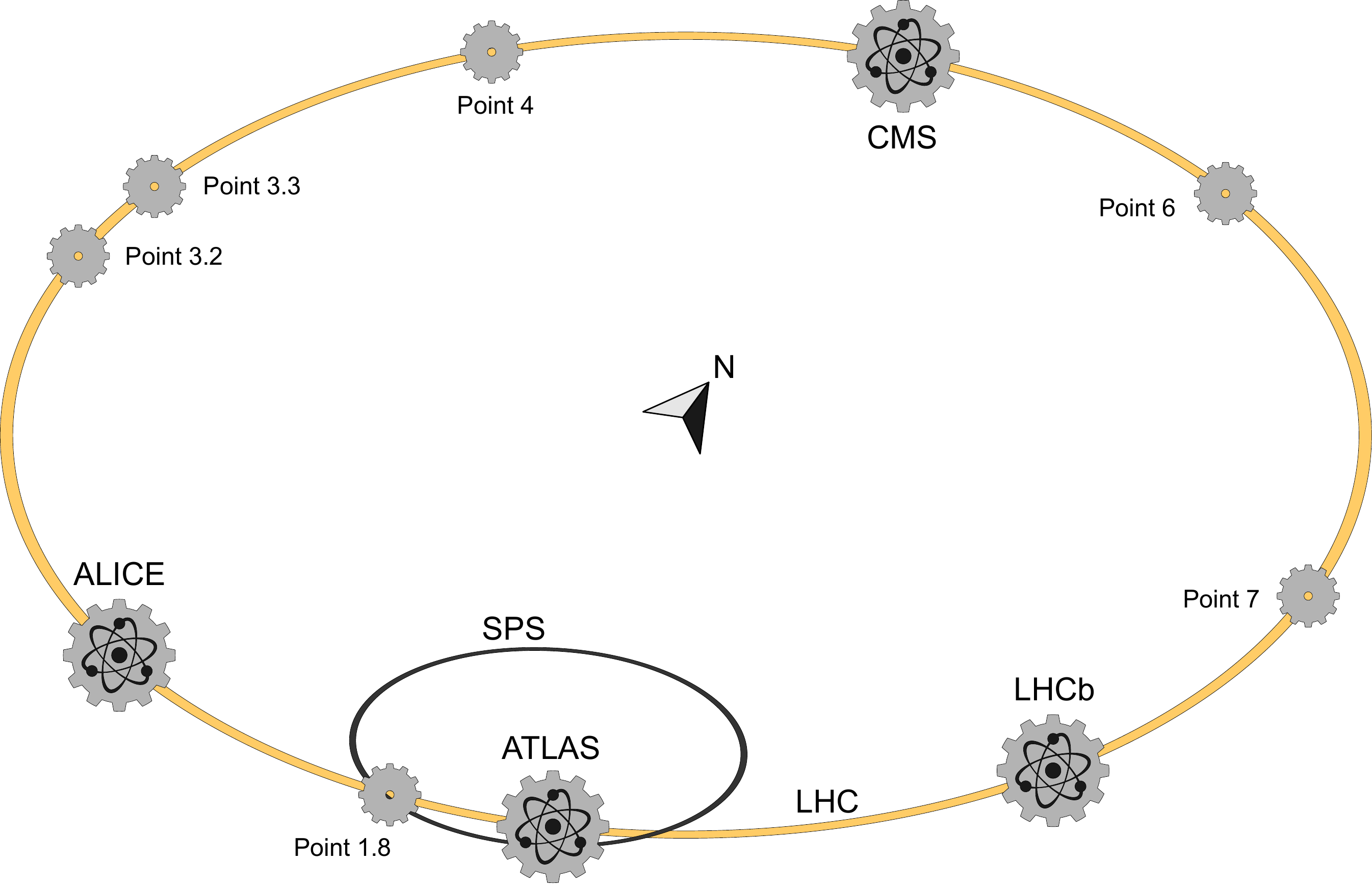}
	\caption{Diagram of the LHC and the related experiments (adapted from~\cite{stancari2015conceptual}, \copyright~2014-2017 CERN, license: \href{https://creativecommons.org/licenses/by/3.0/}{CC-BY-3.0}).} \label{fig:lhc_view}
\end{figure}

\begin{figure}
	\includegraphics[width=\columnwidth]{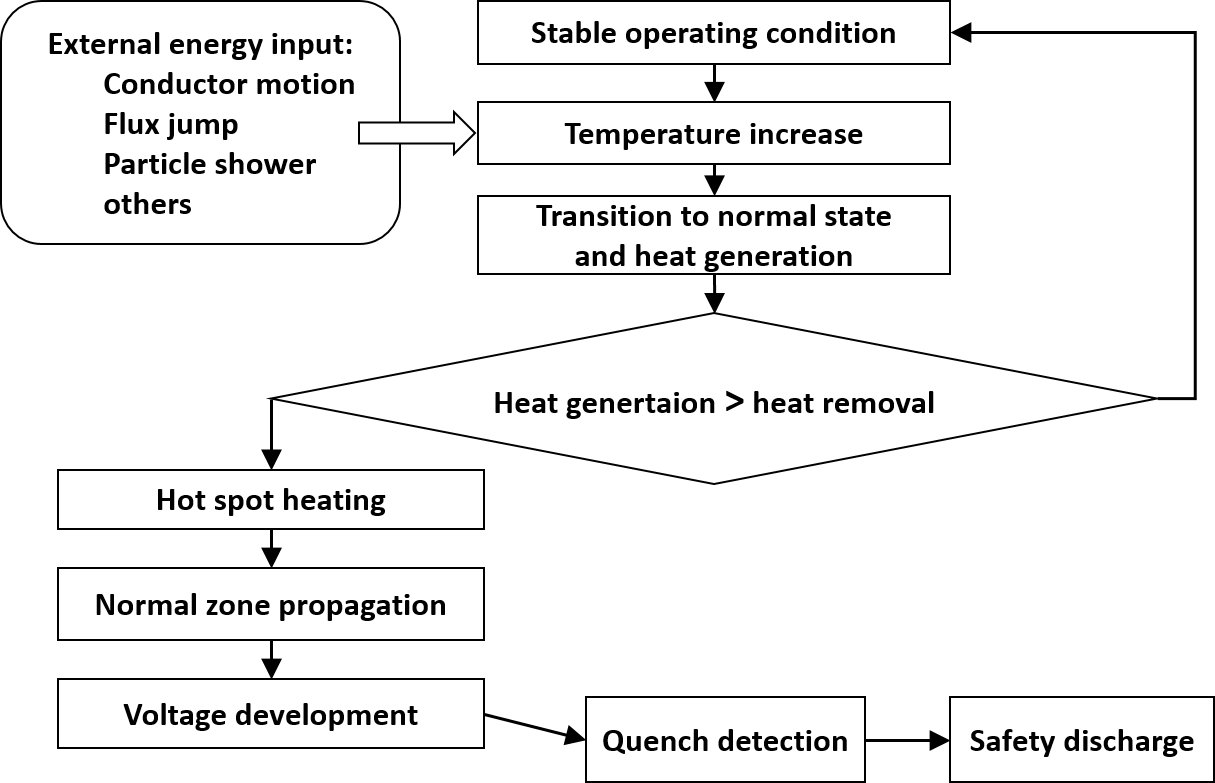}
	\caption{Flowchart describes what typically happens in case of an energy dissipation inside superconducting component (adapted from~\cite{bottura2014cable}, \copyright~2014-2017 CERN, license: \href{https://creativecommons.org/licenses/by/4.0/}{CC-BY-4.0}).} \label{fig:flow_mps}
\end{figure}

\begin{table}
\caption{Nominal conditions in the main dipole circuits of the LHC at the beginning and at the end of ramping up \citep{evans2008lhc}.} \label{tab:lhc_param}
\label{tab:LHClevels}
\centering
\begin{tabular}{lcc}
\toprule
Parameter & Injection & Collision \\
\midrule
Proton energy [\si{\tera\electronvolt}] & \num{0.450} & \num{7} \\
Magnetic field [\si{\tesla}] & \num{0.535} & \num{8.33} \\
Supply current [\si{\ampere}] & \num{763} & \num{11850} \\
Energy stored [\si{\mega\joule}] & \num{4.483} & \num{1081.253} \\
\bottomrule
\end{tabular}
\end{table}

The \gls{LHC}, the largest and most powerful accelerator in the world, is divided into eight sectors (octants) (Fig.~\ref{fig:lhc_view}).
The tunnel itself contains strings of superconducting magnets, accelerating cavities and many other necessary instruments.
Two vacuum beam pipes are going through central part of an iron yoke of magnets.

The particles are produced and initially accelerated by the chain of smaller accelerators.
Then, particles are delivered into the \gls{LHC} with energy at injection level (see Tab.~\ref{tab:lhc_param}).
During every turn around the whole trajectory, the energy of particle raises and synchronously the magnetic file produced by bending magnets must also be increased.
This ramping process takes some time before the machine achieves a condition in which collisions are initiated.
In this state, every \SI{25}{\nano\second} two particle clouds (bunches) collide in four interaction points denoted in the Fig.~\ref{fig:lhc_view} by the names of experiments: \gls{ATLAS}, \gls{CMS}, \gls{ALICE}, and \gls{LHCb}.
Products from each collision are observed by dedicated systems of particle detectors.

The primary goal of whole engineering effort at the \gls{LHC} is to maintain the collision state of the accelerator as long as possible to give a chance to maximize the number of observed events.
However, the quality of beams is decreasing with each collision, and, at some point, they stop being useful for physics experiments.
At this stage of operation, the beams are dumped, the machine must ramp down and be filled with particles again.
This whole work cycle can be interrupted at any time by a malfunction of one of the thousands of elements of the accelerator.

\begin{table}
\centering
\caption{General overview of the circuits powering the superconducting magnets of the LHC \citep{LHC_layout, LHC_quench_db}. The number of quenches as reported on 10 October 2016.} \label{tab:magnets}
\label{tab:LHCcircuits}
\begin{tabular}{lccc}
\toprule
\makecell[l]{LHC\\Circuit} & \makecell{No of\\circuits} & \makecell{No of magnets\\ in one circuit} & \makecell{No of\\quenches} \\
\midrule
RB & \num{8} & \num{154} & \num{1270} \\
RQ  & \num{16} & \num{47} & \num{64} \\
IT & \num{8} & \num{4} & \num{18} \\
IPQ & \num{78} & \num{2} & \num{323} \\
IPD & \num{16} & \num{1} & \num{53} \\
\midrule
\SI{600}{A} EE & \num{202} & $m$ & \multirow{3}{*}{\num{425}} \\
\SI{600}{A} EEc & \num{136} & \num{1} or \num{2} & \\
\SI{600}{A} & \num{72} & \num{1} & \\
\midrule
$\num{80}\div\SI{120}{A}$ & \num{284} & \num{1} & \num{116} \\
\SI{60}{A} & \num{752} & \num{1} & \num{44} \\ 
\bottomrule
\end{tabular}
\flushleft
\footnotesize{RB - Main Dipole; RQ - Main Quadrupole; IT - Inner Triplet; IPQ - Individually Powered Quadrupole; IPD - Individually Powered Dipole; EE - Energy Extraction; EEc - Energy Extraction by crowbar; $m$ - an amount of magnets in circuits is not constant in this class of circuits.} 
\end{table}

There are \num{1232} dipole and \num{392} quadrupole magnets that are crucial elements of the \gls{LHC} (see Tab.~\ref{tab:magnets} for the approximate list).
The coils of those electromagnets are wound up with multi-filament cables.
The filaments are made with niobium-titanium \chem{Nb-Ti} alloy, with a copper matrix surrounding them.
This kind of coils produces a magnetic field of \SI{8}{\tesla}, sufficient to drive particles along the ring at \SI{7}{\tera\electronvolt} energy.
The coil conducts a high superconducting current, but sometimes, locally, in a random and uncontrolled way, it becomes normally-conducting.
This event (a quench) is hazardous because it is connected with burst dissipation of energy stored in the superconducting circuit.
It is not a malfunction, but a physical phenomenon which takes place in any superconducting circuit which does not meet a condition of cryogenic stability \cite{Wilson_book}.
The superconducting magnets applied in the accelerators are designed as not safe in that sense.
Many other design constraints make this an only feasible possibility.
Therefore the \gls{QPS} was created at the \gls{LHC} \cite{denz2006electronic, steckert2017design}.

\Gls{QPS} is a sophisticated subsystem dedicated to magnet coils monitoring and anomaly detection, supervising working condition changes during various phases of the system operation.
The voltages on coils, busbars and current leads are acquired and stored in a database.
The malfunctions or quenches are detected on-line when a value of the voltage exceeds a safety threshold.
When this state lasts longer than a discrimination time, a trigger signal is generated to stop the operation of the whole accelerator and to discharge the energy stored in its circuits safely.
The diagram presented in Fig.~\ref{fig:flow_mps} summarizes the described scenario. Any undetected quench (false negative) can lead to catastrophic damages due to huge energy stored in magnetic field (Tab. \ref{tab:lhc_param}). This energy must be discharged in a controlled manner. 

For the High Luminosity upgrade of the \gls{LHC} (\gls{HL-LHC}) a new generation of niobium-tin \chem{Nb_3Sn} superconducting magnets will be installed as the inner triplets quadrupoles (low beta quadrupoles MQXFA/B) and the \SI{11}{\tesla} dipoles (MBH) \cite{11Tstatus2015}. In particular, the inner triplets of the points 1 and 5, \gls{ATLAS} and \gls{CMS}, will be replaced while the \SI{11}{\tesla} dipoles will take the place of the standard \gls{LHC} main dipoles on both the sides of point 7. The new magnets will be fed by a superconducting link with niobium-boron \chem{MgB_2} cables \cite{ballarino2014link}.

The technologies used for reaching high magnetic field also implicate the development of a new protection system, and in particular of a dedicated detection system \cite{denz2017next}. In fact, this kind of coils suffers from not only a quench but also from the so-called flux jumps \cite{11Tquench2014}, affecting the nature of voltage waveforms describing the state of superconducting magnets. In the ongoing magnet tests, the occurrences of these with the relate voltage spikes at a low current rate ($I < \SI{4}{\kilo\ampere}$) could represent an issue for the classical detection parameters (voltage threshold of $\pm \SI{100}{\milli\volt}$ and the evaluation time of \SI{10}{\milli\second}). As a consequence, the strategy for the quench detection should be replaced with a new one based on dynamically set detection parameters \cite{denz2017next}. 
\subsection{Source of data for this study}
\label{subsec:lhc:data_orig}

\begin{figure*}
\centering
\includegraphics[width=0.85\textwidth]{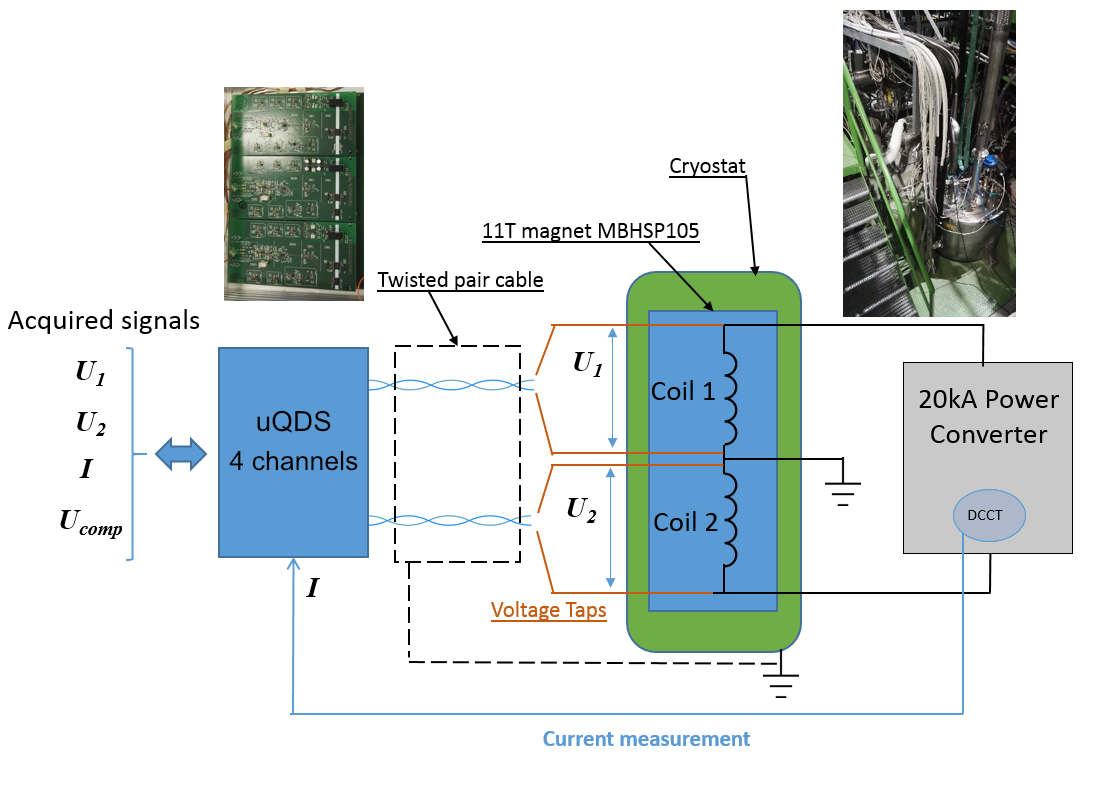}
\caption{Measurement setup for the \SI{11}{\tesla} MBHSP105 magnet tests.} \label{fig:lhc:meas_setup}
\end{figure*}

Any superconducting circuit which does not meet a condition of cryogenic stability needs permanent monitoring (logging) during testing, commissioning, and operation. 
In case of any event (e.g., quench) in a superconducting circuit a data with higher time resolution is also acquired and stored for analysis. 
Therefore, two dedicated database services were built at \gls{CERN}.
The logging data is stored in \gls{CALS}, whereas the data acquired during an event is kept in a separate system named \gls{PMS}. 

The data used in this study is of the logging type.
It was acquired during testing a new type of magnet designed for the \gls{HL-LHC} project.
The test was conducted at the Superconducting Magnet Test Facility (SM18) in November 2016. 
The test was performed on single aperture dipole magnet dedicated to delivering magnetic filed on the level of \SI{11}{\tesla}. 
Therefore the coils of this magnet were wound with a cable made of niobium-tin \chem{Nb_3Sn} superconducting material. 
The exact designation of the magnet is MBHSP105 \cite{11Tstatus2015}.

The goal of the test was to train the magnet. 
A magnet training is an iterative procedure of magnet powering.
At first, during ramping up a current, a magnet loses superconducting state (quench) long before reaching the expected critical current.
During the next attempt, the current that could be reached before quench is higher.
The process continues over all the next attempts, and the maximum current that could be reached increases quench after quench, slowly approaching a plateau.

In our case, the plateau was reached after several runs.
Each run begun with ramping of magnet current with rate \SI[per-mode=symbol]{50}{\ampere\per\second}. 
After reaching the level of \SI{8}{\kilo\ampere}, the rate was lowered to the value of \SI[per-mode=symbol]{10}{\ampere\per\second}. The ramping with this rate was maintained up to a quench. 

During the runs, the current and voltages in the coils of the magnet were measured using a device named \gls{uQDS} (ver. 1.0) \cite{denz2017next}.
The measurement setup is shown in the Fig.~\ref{fig:lhc:meas_setup}.
The device was built with an analog to digital converter \gls{ADC} of the \gls{SAR} type with \SI{20}{\bit} resolution. 
The digital control of the acquisition was built with \gls{FPGA} circuit of the IGLOO$^{\scriptscriptstyle\textregistered}$2  type from Microsemi$^{\scriptscriptstyle\textregistered}$. 
The sampling rate was \SI{9.3}{\kilo\SamplePerSecond} (sampling period \SI{107.5}{\micro\second}). 

The data of \gls{PM} kind was also acquired during this training with a sampling rate of \SI{100}{\kilo\SamplePerSecond} but it was not used in this study.

\section{State of the art}
\label{section:state-of-the-art}

\subsection{Overview}
\label{subsection:state-of-the-art:overview}

Monitoring time series signals changes is critical in many areas of engineering and real-life applications.
It is mostly because roughly \SI{80}{\percent} of the signals which occur in the world are temporal in their nature. 
Consequently, anomaly detection was heavily explored as a field over the last several decades and many methods were developed to address this challenging task \cite{Chandola2008Comperative, Chandola2009Anomaly, Pimentel2014Novelty}.
It is worth noting that most of the anomaly detection tasks deal with quite asymmetric datasets which means that there are far few cases of anomalous behavior than regular ones.
Furthermore, labeling the data is challenging task.
Characteristics mentioned above lead to the preference of unsupervised methods over supervised ones when it comes to real-life applications.
An ideal anomaly detection system should: 
\begin{itemize}
 \item be able to detect anomalies with the highest possible accuracy,
 \item be trained in unsupervised fashion,
 \item trigger no false alarms,
 \item work with data in a real-time,
 \item be completely adaptable (no hyper-parameters tuning).
\end{itemize}

Unfortunately, it is difficult to construct such a system not only because of a challenge to meet all of the requirements at the same time but also for the sake of a data profile.
For instance, how frequently a model should be updated to account for seasonality of changes in the data is not always clear. 
Furthermore, real-time performance is not always at the premium.
However, due to the rise of data volume and an increasing demand for speed at which the system should deliver results, we may expect the growing demand for real-time performance.

Anomaly detection systems in real-life applications are not ideal which means that they do not meet all the requirements enumerated on the list.
They do not have to, as very often it is enough that a system detects most of the anomalies in a reasonably short time.
Sometimes, however, for the sake of a task profile, it is critical that a system does not generate false alarms, even at the expense of a slightly lower overall accuracy.
In some other cases, response accuracy is not as important as a low response time of the system.
It can be observed while analyzing how anomaly detection systems developed over the past few decades \cite{Pimentel2014Novelty} that there is a trade-off between response accuracy and reaction time.
Consequently, depending on an expected performance three different groups of anomaly detection systems may be distinguished:
\begin{itemize}
 \item offline,
 \item partially online,
 \item online.
\end{itemize}

The first category of the systems operates in an offline fashion which means that they are trained offline and work offline.
Such solutions are well suited for processing large volumes of data at relatively low pace and usually require access to the whole dataset.
Examples of such systems used in industrial applications are \gls{EGADS} developed by Yahoo \cite{Laptev2015Generic} and \gls{RPCA} \cite{Candes2009Robust}.
The \gls{EGADS} is based on the assumption that integration of several methods within a single framework helps to address different kinds of anomalies.
Such an approach is very sound in principle but comes at the cost of processing time which results from a necessity of weighting and incorporating contributions of different methods.
There are also other solutions which can be classified into a category of offline systems \cite{Keogh2005Hot, Akouemo2016Probabilistic}.
  
The second category of the algorithms is trained offline and work online.
Complex and large models usually need to be trained offline because of the time it takes to complete the process.
However, sometimes the model is small enough to be trained online, but it is still beneficial to conduct the training process offline as a result of an application profile and the system requirements.

In some applications, a system should be more sensitive to seasonal changes, and it is essential to train a model only during specific periods of time.
There is a whole branch of clustering-based algorithms which are trained offline to subsequently work online \cite{Zengyou2002Squeezer, Jaing2001Two, He2003Disovering, Duan2009Cluster, Lee2011Self, Ahmed2007Multivariate, Faria2013Novelty, Spinosa2007Olindda}.
Time series are clustered according to their properties, and all the outliers which do not belong to one of the clusters are considered anomalies.
An amount of clusters and the classification threshold are two of the more critical parameters which are to be chosen for the applications of those algorithms.

One of the standard solutions which fall in the second category is an approach based on \gls{OC-SVM}.
Several implementations of anomaly detection systems based on \gls{OC-SVM} were proposed, and promising results were reported \cite{Ma2003Time, Zhang2007One, Su2015Anomaly, s141120713, 4463657, scholkopf2000support, Scholkopf:2001:ESH:1119748.1119749}.
The methods based on \glspl{RNN}, described in \ref{subsection:state-of-the-art:rnns}, also belong to that category.

The third kind of the anomaly detection systems adapts online which means that all the novelties which are detected are incorporated in the model \cite{Hole2016Anomaly, wielgosz2017using}.
Next time the same phenomenon occurs in the input signal to the system, it will not be considered as an anomaly.
In such an approach, the system continually adapts to the changing environment which may be beneficial in many scenarios.
However, there are applications in which due to the seasonality it is recommended to update a model in well-defined moments of time. 

There exist more advanced streaming anomaly detection methods, such as ESD, \gls{ARIMA}, and Holt-Winters \cite{Wang2011Statistical, Angelov2014Anomaly, Bianco2001Outlier, Ekberg2011Network}, which are used in many industrial applications.
A broad analysis of the modern anomaly detection systems is beyond the scope of this paper, for more in-depth review please see  \cite{Chandola2009Anomaly, Chandola2008Comperative, Pimentel2014Novelty}.
It is worth emphasizing that the area of novelty detection is expanding very fast which is driven by an exponential growth of available information and rising need for knowledge extraction.
We may expect this trend to intensify as a result of an introduction of new hardware platforms which are capable of processing data faster \cite{TPUgoogle2017, catapult2011, MS_FPGA_2016}. 

\subsection{\glsentrylongpl{RNN}}
\label{subsection:state-of-the-art:rnns}

The \Glspl{RNN} fundamentally differ from the \glspl{FNN}.
The recurrent neural models learn to recognize patterns in the time domain.
Consequently, they are capable of modeling signals in which patterns span over many time steps.

For many years, factors limiting the development and engineering applicability of recurrent neural networks existed.
Those restrictions were associated with the possibility of learning very long patterns, with sequences length in tens or hundreds.
Classical \glspl{RNN} were not able to learn them due to the so-called vanishing (or exploding) gradient phenomenon.
Scientists working in an \gls{RNN} domain were aware of it occurring, not only in \glspl{RNN} but also in deep \glspl{FNN}.
Therefore, extensive research was conducted, and in 1997 it resulted in development of the \gls{LSTM} architecture by J\"urgen Schmidhuber \citep{hochreiter1997long}.
Unfortunately, due to the lack of computing power and limited available data quantities, \gls{LSTM}-type networks were developing slowly.
In the recent years, however, there was a considerable progress in \glspl{RNN}.
Many variants of the original \gls{LSTM} algorithm were introduced, optimizing the original architecture.
One of such a modifications is \gls{GRU} \citep{chung2015gated, chung2014empirical}.
Short presentation of \gls{GRU} architecture can be found in ~\ref{appx:gru}.

\Gls{RNN} networks were employed for anomaly detection task in different operational setups \cite{malhotra2015long, bontemps2016collective, nanduri2016anomaly, marchi2015novel, marchi2015nonlinear, chong2017abnormal}.
The authors of \cite{malhotra2015long} proposed an architecture of \gls{LSTM}-based anomaly detector which incorporates both hierarchical approach and multi-step analysis.
The proposed model capitalizes on a property of generalization which results from stacking of several \gls{RNN} layers.
The system in \cite{malhotra2015long} was trained on regular data and verified on data containing anomalies.
The authors used Gaussian distribution of an error signal - the difference between predicted and real values.
Consequently, the module predicts several time steps into the future and fits the estimated error into the Gaussian distribution.
Four different datasets were used to demonstrate the performance of the proposed solution.

There is a whole branch of \gls{LSTM}-based anomaly detectors which exploit a property of inconsistent signal reconstruction in the presence of anomalies \cite{marchi2015novel, marchi2015nonlinear}.
The authors trained the model of the autoencoder on regular data and set a threshold above which the reconstruction error is considered an anomaly.
The papers deal with acoustic signals, but such an approach may be efficiently employed in other domain such as videos \cite{chong2017abnormal}.
Systems based on those principles may be trained in an end-to-end fashion. 
The presented models \cite{marchi2015novel, marchi2015nonlinear} were trained on the publicly available datasets, and the results are superior in comparison with the other solutions. 

\section{Quantization algorithm}
\label{section:grid}

\subsection{Previous work}
\label{subsection:grid:previous-work}

\begin{table*}
\caption{Notation used in quantizaton equations (\ref{eq:input_norm}) - (\ref{eq:edges}). \label{tab:grid:notation}}
\centering
\begin{tabular}{cl}
\toprule
$n$ & number of samples \\
\midrule
$m$ & number of classes (categories, bins); $m \in \mathbb{N}_{>0}$ \\
\midrule
$S_{\mathit{in}}$ & signal input space \\
\midrule
$S_{\mathit{norm}}$ & normalized input space \\
\midrule
$S_{\mathit{qs}}$ & signal space after static quantization \\
\midrule
$S_{\mathit{qa}}$ & signal space after adaptive quantization \\
\midrule
$\vec{edges}_i$ & $i$-th quantization edge, see (\ref{eq:edges}) \\
\midrule
$\vec{sorted\_samples}_i$ & \makecell[l]{$i$-th sample in the ascending sorted array of all\\available signal samples} \\
\bottomrule
\end{tabular}
\end{table*}

\begin{table*}
    \caption{Percentage of samples per single class in training set. \label{tab:cardinality}}
    \centering
    \begin{tabular}{lcccc}
        \toprule
        & & & \multicolumn{2}{c}{\si{\percent} of total samples per class} \\
        \cmidrule{4-5}
        & no of classes & median bin width & minimum & maximum \\
        \midrule
        \multirow{3}{*}{static grid} & \num{8} & \num{0.125} & \SI{0}{\percent} & \SI{\sim 100}{\percent} \\
        & \num{16} & \num{0.0625} & \SI{0}{\percent} & \SI{\sim 100}{\percent} \\
        & \num{100} & \num{0.01} & \SI{0}{\percent} & \SI{\sim 81}{\percent} \\
        \midrule
        \multirow{3}{*}{adaptive grid} & \num{8} & \num{0.00035} & \SI{\sim 8}{\percent} & \SI{\sim 14}{\percent} \\
        & \num{16} & \num{0.00018} & \SI{\sim 2}{\percent} & \SI{\sim 7}{\percent} \\
        & \num{100} & \num{0.00003} & \SI{\sim 0}{\percent} & \SI{\sim 1}{\percent} \\
        \bottomrule 
    \end{tabular}
\end{table*}

\begin{figure*}
    \centering
    \begin{subfigure}{0.33\hsize}
         \hspace{-0.1\hsize}
         \includegraphics[trim={0 0 405pt 0},clip,width=\hsize]{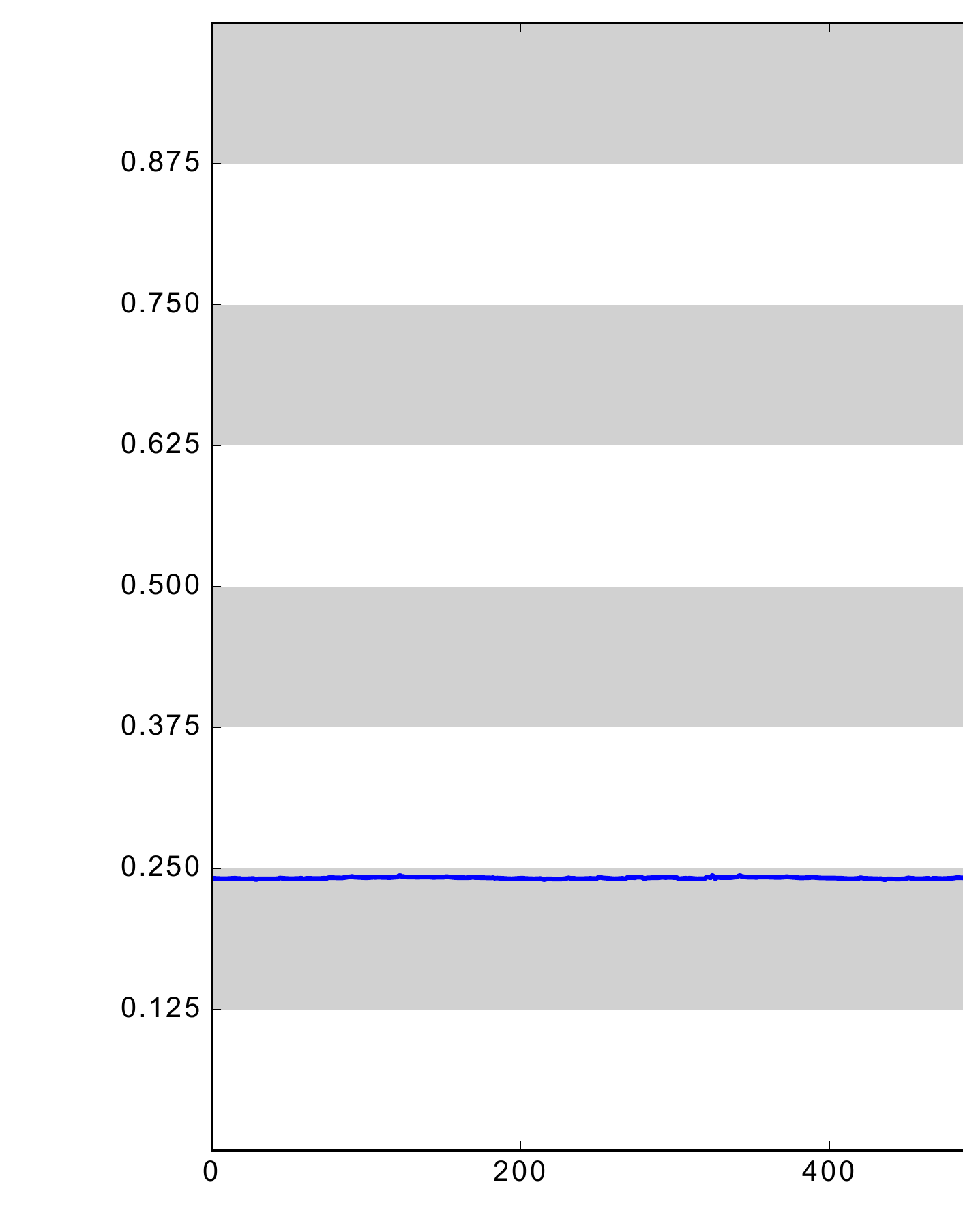}
         \caption{Static grid \label{subfig:grid:static_8}}
    \end{subfigure}%
    \begin{subfigure}{0.33\hsize}
         \hspace{-0.1\hsize}
         \includegraphics[trim={0 0 405pt 0},clip,width=\hsize]{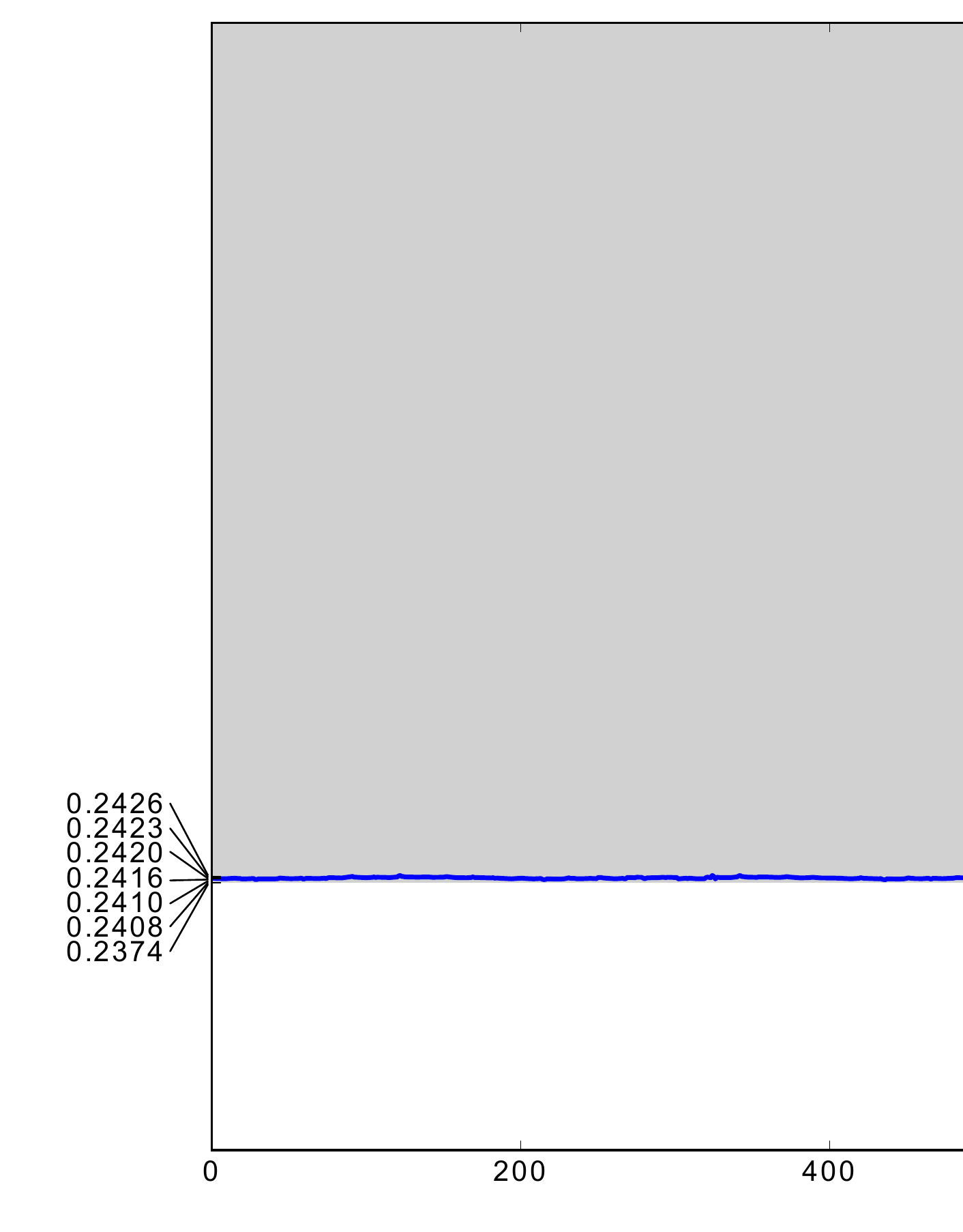}
         \caption{Adaptive grid \label{subfig:grid:adaptive_8}}
    \end{subfigure}%
    \begin{subfigure}{0.33\hsize}
         \hspace{-0.1\hsize}
         \includegraphics[trim={0 0 405pt 0},clip,width=\hsize]{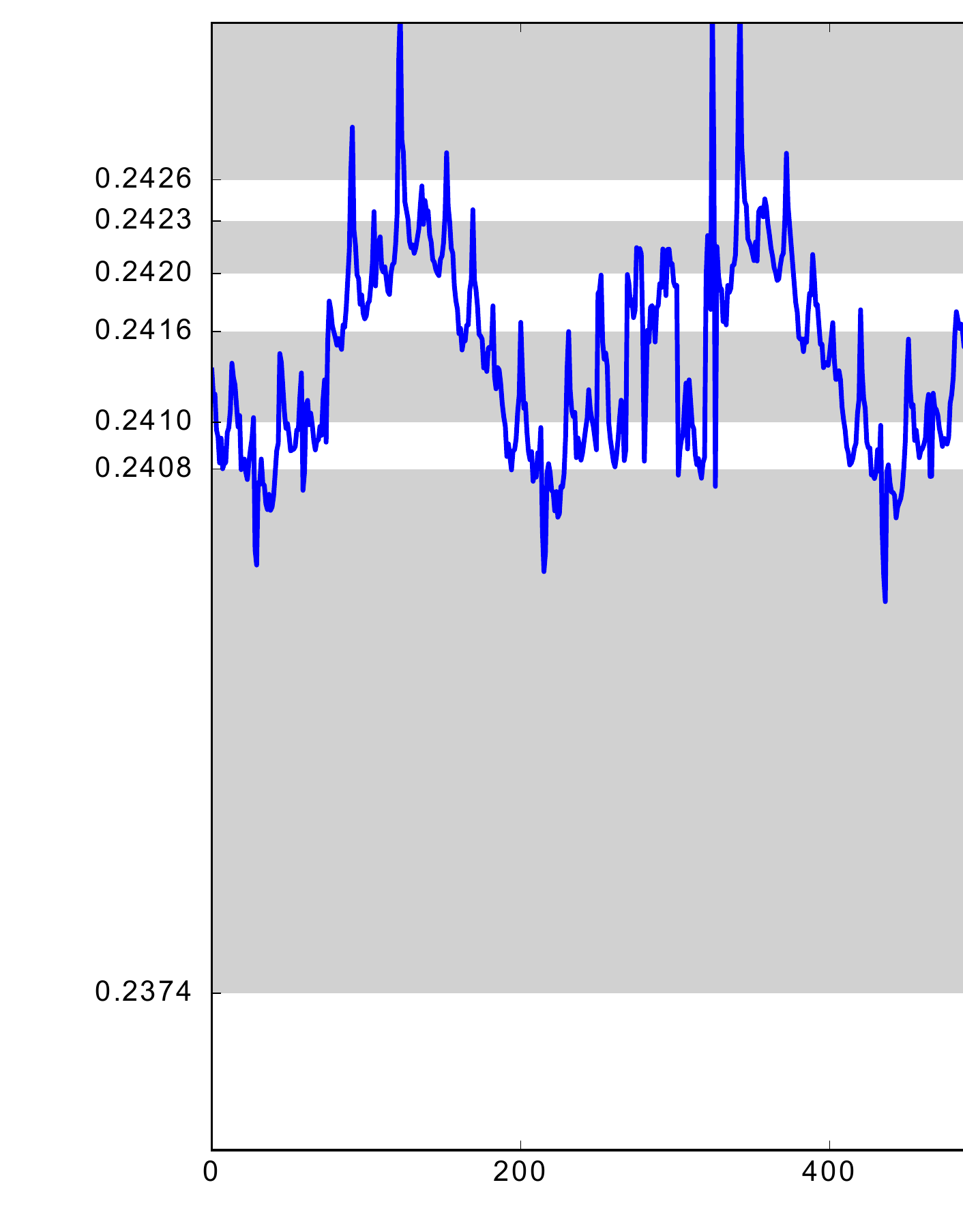}
         \caption{Zoom of adaptive grid \label{subfig:grid:adaptive_8_zoom}}
    \end{subfigure}
    \caption{Static (\subref{subfig:grid:static_8}) vs. adaptive (\subref{subfig:grid:adaptive_8}, \subref{subfig:grid:adaptive_8_zoom}) quantization; $m=8$. \label{fig:grid}}
\end{figure*}

\begin{figure*}
    \centering
    \begin{subfigure}{\hsize}
        \centering
        \includegraphics[width=0.8\hsize]{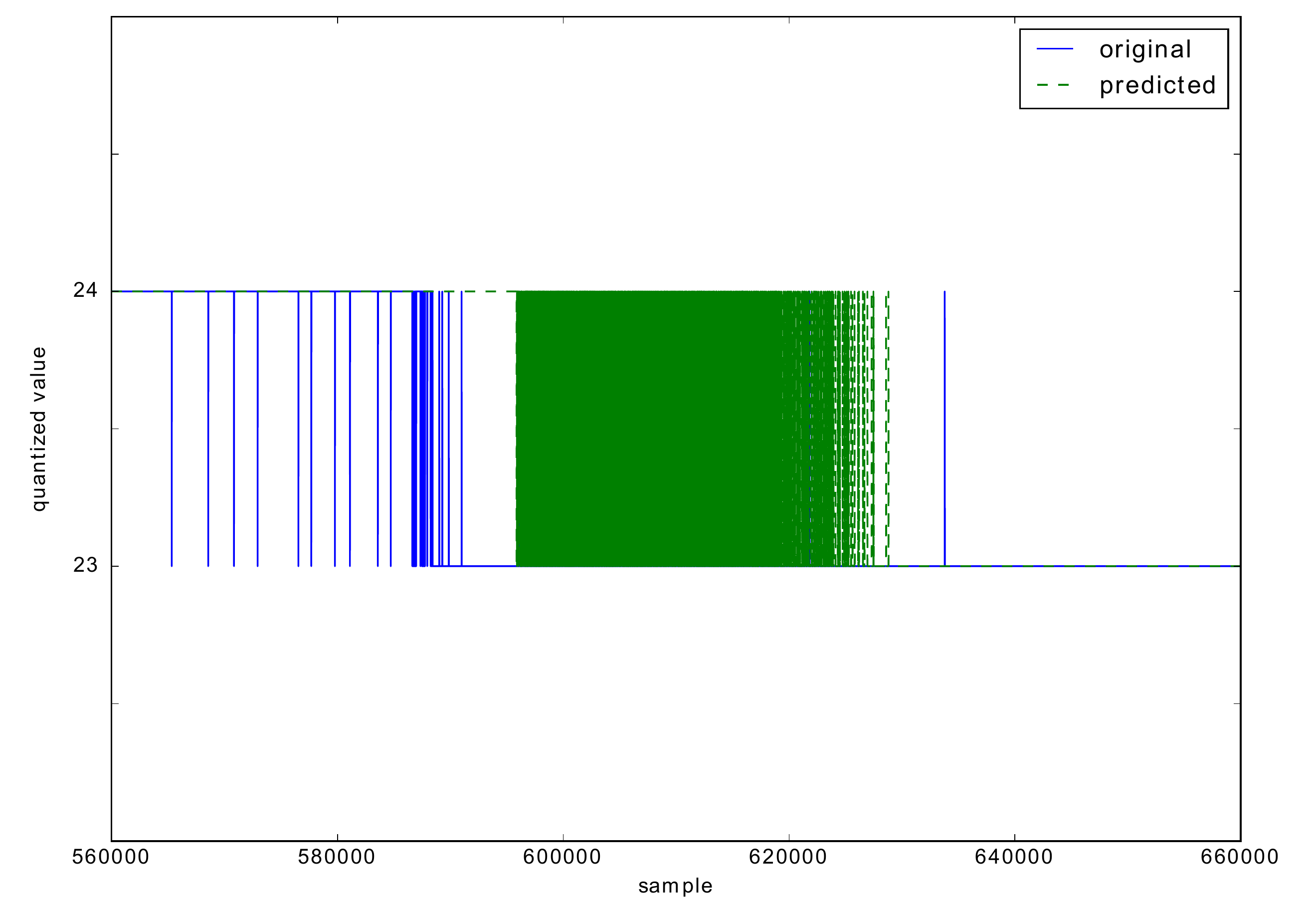}
        \caption{Static quantization; $m=100$ \label{subfig:shift:static}}
    \end{subfigure}
    \begin{subfigure}{\hsize}
        \centering
        \includegraphics[width=0.8\hsize]{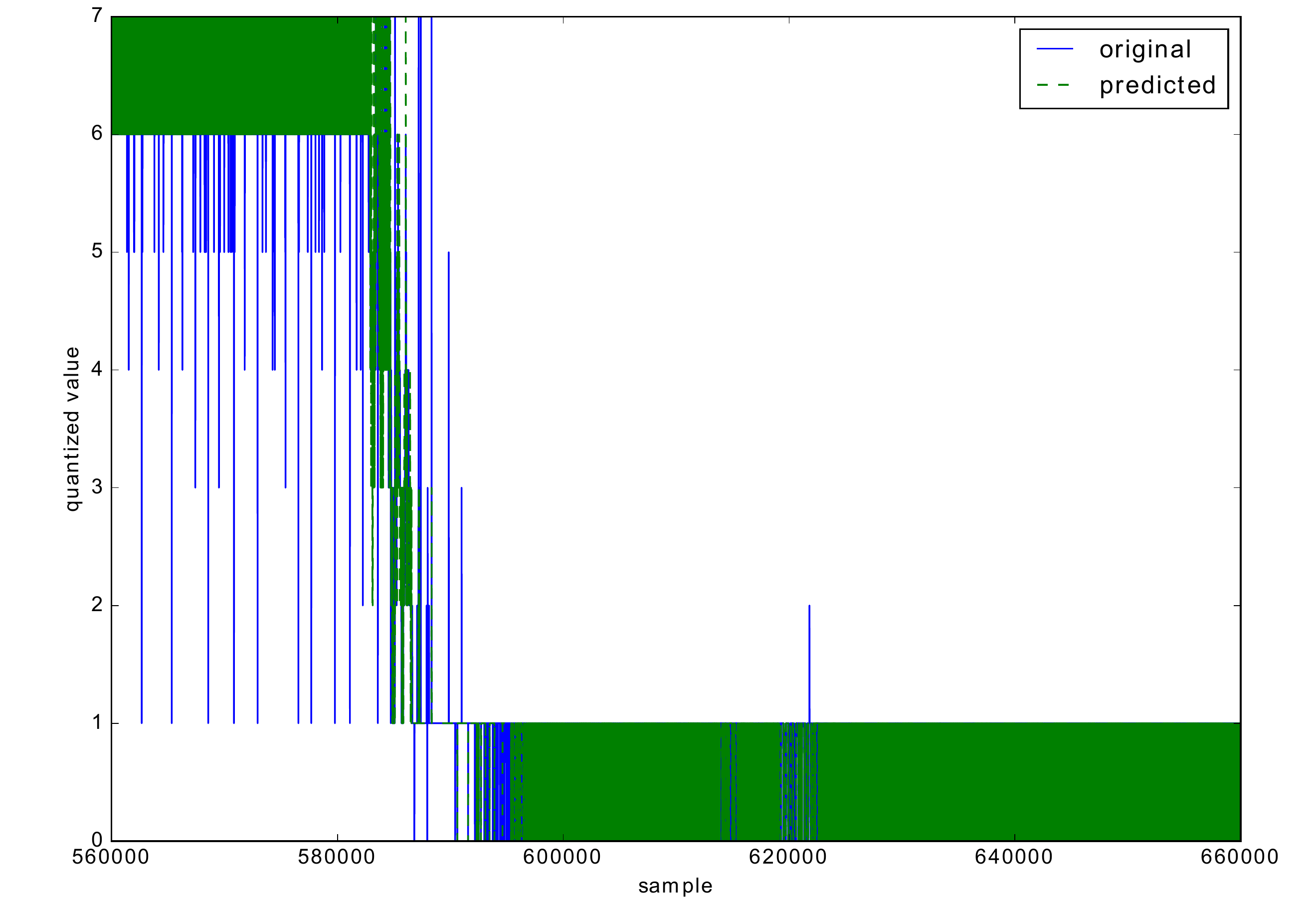}
        \caption{Adaptive quantization; $m=8$ \label{subfig:shift:adaptive}}
    \end{subfigure}
    \caption{Models reacting to vertical shift in signal (same data fragment). \label{fig:shift}}
\end{figure*}

In the authors' previous work concerning superconducting magnets monitoring \citep{wielgosz2017lstm} a \gls{RMSE} approach was used.
It showed that \glspl{RNN} are in fact able to model magnets behavior.
However, it has several drawbacks that make it hard to use in practical anomaly detection applications.

Firstly, to adequately analyze anomalies using the \gls{RMSE} it would be necessary to select a detection threshold.
Such a threshold would be very arbitrary, since it is hard to discern what value, allowing to detect all anomalies, would be appropriate based on the results obtained from all the data, including regular operation.

Secondly, the resolution of such an anomaly detection would depend on used window size.
Additionally, choosing too broad a window would result in anomaly potentially 'drowning' in the correct data, while choosing too small could result in false positives.
The window size would also influence the trained \gls{RNN} accuracy.

Described drawbacks resulted in authors' decision to switch from regression to a form of predictive classification. 
Initially, the signals were converted to classes using a static, evenly-spaced grid, mapping whole signal amplitude \citep{wielgosz2017recurrent}.
Both input signals and output one were mapped, and the model task was to predict output category given a tensor of input classes correctly.
When the prediction and real signal did not match over a certain amount of samples, an anomaly was reported.

A static quantization process is mapping signal input space $S_{\mathit{in}}$ to $m$ classes (see Tab.~\ref{tab:grid:notation} for notation used), that can potentially be represented by $\ceil{\log_2 m}$ bits instead of initial \num{32} or {64} per value.
At first, as given by (\ref{eq:input_norm}) - (\ref{eq:input_norm_mapping}), the signal is normalized.
Next, the normalized signal values are mapped to categories, using $m$ evenly-spaced bins spanning whole signal amplitude (\ref{eq:norm_qs}) - (\ref{eq:norm_qs_mapping}).

\begin{equation}
    S_{\mathit{in}} : \mathbb{R}^{1 \times n} \xRightarrow{\Pi_{\mathit{norm}}} S_{\mathit{norm}} : \left\lbrace 0 \ldots 1 \right\rbrace ^ {1 \times n},
\label{eq:input_norm}
\end{equation}

\begin{equation}
    \Pi_{\mathit{norm}} : \bigwedge_{x \in S_{\mathit{in}}} \bigvee_{y \in S_{\mathit{norm}}} ~ y = \frac{x - \min S_{\mathit{in}}}{\left| \max S_{\mathit{in}} - \min S_{\mathit{in}} \right|}.
\label{eq:input_norm_mapping}
\end{equation}

\begin{equation}
    S_{\mathit{norm}} \xRightarrow{\Pi_{\mathit{qs}}(m)} S_{\mathit{qs}} : \left\lbrace 0 \ldots m-1 \right\rbrace ^ {1 \times n},
\label{eq:norm_qs}
\end{equation}

\begin{equation}
    \Pi_{\mathit{qs}}(m) : \bigwedge_{x \in S_{\mathit{norm}}} \bigvee_{y \in S_{\mathit{qs}}} ~ y = 
\begin{cases}
y \leqslant x \cdot m < y+1 \text{~if~}  x < 1 \\
y = m-1 \text{~if~}  x=1
\end{cases}
.
\label{eq:norm_qs_mapping}
\end{equation}

As a result of conducted experiments analysis, as well as formal static quantization algorithm scrutiny, the authors concluded that using evenly-spaced grid will not allow detecting anomalies effectively.
It was due to the algorithm mapping most of the data to a minuscule number of categories, with the majority of them almost never being used (see Fig.~\ref{subfig:grid:static_8} and Tab.~\ref{tab:cardinality}).
The end effect was that, on the one hand, it took a long time for a model to adapt to a vertical shift in data (Fig.~\ref{subfig:shift:static}), resulting in false anomalies, and, on the other hand, smaller anomalies would have no chance to be detected.
Additionally, since most of the categories were barely used, it resulted in wasting resources.

\subsection{Adaptive grid}
\label{subsection:grid:adaptive}

Analysis of previously mentioned experiments and algorithms resulted in a conclusion that a more advanced algorithm is needed.
It should avoid the threshold-selection problem, allow to harvest the \glspl{RNN} potential by using classification instead of regression and optimally use available resources.
As a consequence, adaptive grid quantization algorithm was developed. Its principle of operation is mapping the input space to a fixed number of categories (bins) in such a way, that all categories have (ideally) the same samples cardinality as it is described by equations (\ref{eq:norm_qa}) - (\ref{eq:edges}).
As a result, bins widths are uneven, explicitly adjusted to the input signal (see Fig.~\ref{subfig:grid:adaptive_8_zoom} and Tab.~\ref{tab:cardinality}). 
Each of the signals used in the model training has its own bins edges calculated.
This approach allows to potentially maximize the utilization of the grid and minimize the consumption of resources.

\begin{equation}
    S_{\mathit{norm}} \xRightarrow{\Pi_{\mathit{qa}}(m)} S_{\mathit{qa}} : \left\lbrace 0 \ldots m-1 \right\rbrace ^ {1 \times n},
\label{eq:norm_qa}
\end{equation}

\begin{equation}
    \Pi_{\mathit{qa}}(m) : \bigwedge_{x \in S_{\mathit{norm}}} \bigvee_{y \in S_{\mathit{qa}}}
    y = 
\begin{cases}
\!\begin{multlined}
\vec{edges}_y \leqslant x \cdot m < \vec{edges}_{y+1}\\[-10pt] \text{if~}  x < 1
\end{multlined} \\
y = m-1 \text{~if~}  x=1
\end{cases}
,
\label{eq:norm_qa_mapping}
\end{equation}

\begin{equation}
    \vec{edges}: \bigwedge_{0 \leqslant y \leqslant m} \vec{edges}_y = 
\begin{cases}
0  \text{~if~}  y = 0 \\
\!\begin{multlined}
\vec{sorted\_samples}_{y \cdot \ceil{\frac{n}{m}}} \\[-10pt] \text{if~}  0 < y < m
\end{multlined} \\
1  \text{~if~}  y = m
\end{cases}
.
\label{eq:edges}
\end{equation}
\section{System description}
\label{section:system-desc}

\subsection{Principle of operation}
\label{subsection:system-desc:principle}

\begin{figure}
    \centering
    \includegraphics[width=\hsize]{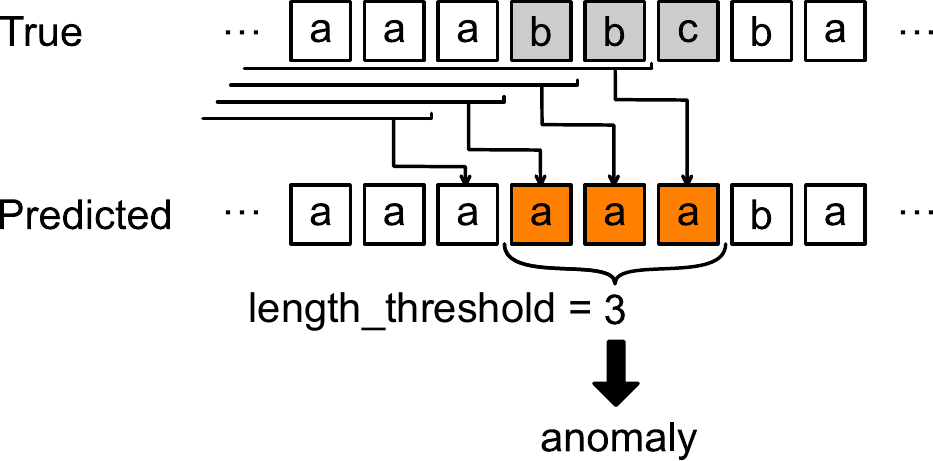}
    \caption{System principle of operation. \label{fig:system-desc:principle}}
\end{figure}

The detector principle of operation is a comparison between predicted and real signal values (Fig.~\ref{fig:system-desc:principle}).
Whenever a new sample arrives, it can be used (in conjunction with previous samples) to predict the category of the next sample.
Assuming that the model was trained to anticipate normal operating conditions ideally, any difference between the prediction and an actual arriving sample category means that an anomaly occurred.

In the practical applications, however, achieving the model perfection is not feasible: the data used to train the model, as well as real samples that the predictions will be compared with, contains noise.
Given large enough pool of samples to learn from, the model should start to predict nearly ideal normal operation values, but even the actual normal operation samples will differ from that ideal due to noise.
When an anomaly occurs, those differences should be much more pronounced.
As a result, a method to discriminate between 'noise anomalies' and actual anomalies is needed.

The simplest discriminating method is to check an anomaly candidate length.
In previous work \citep{wielgosz2017recurrent}, authors assumed that a gap between available history data and prediction (\var{look\_ahead}) must be bigger than \var{length\_threshold}, so that, in case of an anomaly occurring, the model prediction would not get distorted by an irregular input.
The size of this gap, in turn, further affected the model accuracy.
However, after further research on the \gls{RNN} behavior, the authors concluded that this condition is unnecessary since the model should (up to some point) ignore the anomalous sample in favor of available normal operation historical data, with this 'smoothing' capability increasing with \var{look\_back} (history window) length.
Predicting only one step forward (\var[1]{look\_ahead}) has an additional advantage of potentially decreasing system reaction time, especially in conjunction with more advanced anomaly discrimination methods.

\subsection{Setup overview}
\label{subsection:system-desc:overview}

\begin{figure}
    \centering
    \includegraphics[width=0.8\hsize]{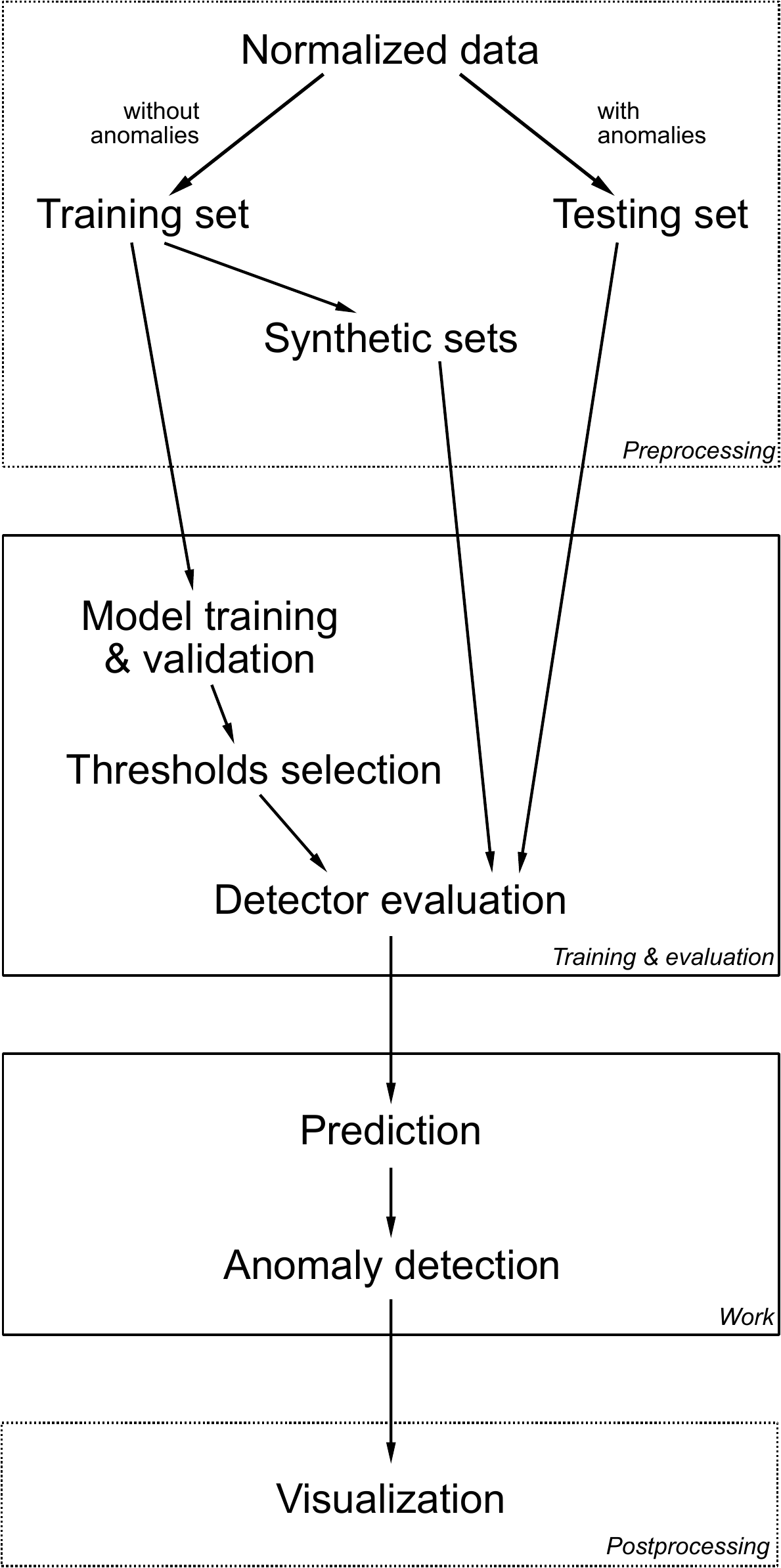}
    \caption{Setup life cycle conceptual overview. \label{fig:system-desc:setup_lifecycle}}
\end{figure}

\begin{figure}
    \centering
    \includegraphics[width=0.8\hsize]{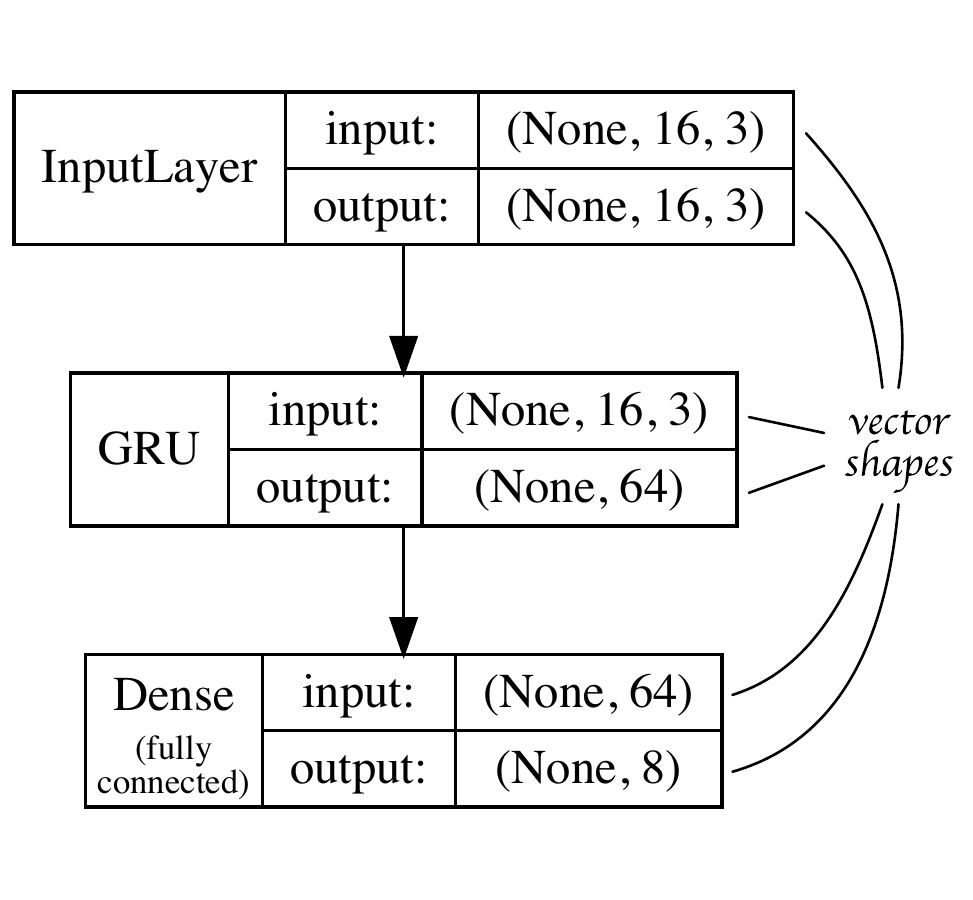}
    \caption{Example model generated for \var{out\_grid}{=8}, \var{cells}{=64}, \var{look\_back}{=16}. \label{fig:system-desc:model}}
\end{figure}

The system is coded in Python, using Keras \citep{chollet2015keras} library with Theano \citep{theano2016} backend for the classifier implementation.
The \obj{detector} module consists of two main sub-modules, \obj{model} and \obj{analyzer}, and a few helper scripts.
The conceptual overview of the single \obj{detector} setup is presented in Fig.~\ref{fig:system-desc:setup_lifecycle}.

The system is prepared to work with normalized data, which are prepared and saved beforehand.
Normalization process takes into account all available data, both from training and testing sets.
All the data fed into the system at a later time would need to be prepared using the same scale, with out-of-range values clipped to \num{0.0} or \num{1.0} as relevant.
All setup variants use the same normalized data.

Depending on the configuration, a particular system setup variant is created.
The configuration includes options of data pre-processing (e.g., number of input/output categories or the bins edges calculation algorithm), the \obj{model} hyper-parameters (such as an amount of layers, number of cells per layer, and batch size) and the \obj{analyzer} rules (like minimum anomaly length or cumulative amplitude).
Most of the configuration options are specified as arrays, allowing to test several setups and compare them easily.
Models trained during each of the setups are automatically saved.
When the \obj{model} with high enough performance is found, it can be loaded and used to test various \obj{analyzer} setups further.

The data quantization is controlled by \var{in\_grid}, \var{out\_grid}, \var{in\_algorithm} and \var{out\_algorithm} configuration parameters. The \var{in\_grid} and \var{out\_grid} control the number of classes for input and output signals, respectively.
At the moment, each of the input channels is quantized using same grid/algorithm combination, analogically for output channels. 
Available algorithms are, as described in section \ref{section:grid}, \textit{static} (\ref{eq:norm_qs_mapping}) and \textit{adaptive} (\ref{eq:norm_qa_mapping}) - (\ref{eq:edges}).

The \obj{model} is the \obj{detector} core.
It is an abstraction layer over the actual classifier.
In the current implementation, the classifier comprises of a configurable number of \gls{GRU} layers from Keras library, followed by a fully connected layer with dimensionality matching \var{out\_grid} parameter value (see Fig.~\ref{fig:system-desc:model}).
However, as long as this abstract interface is preserved, any classifier capable of prediction can be used.
The fitted \obj{model} accuracy should be high enough that when the \obj{detector} setup is tested using normal operation data, it ideally should not report any anomalies (no false positives).

The \obj{analyzer} module uses fitted \obj{model} to generate predictions and compare them with real quantized output signal values.
Whenever it encounters an anomaly candidate (a discrepancy between real and predicted value), it runs a series of checks, according to configured rules, to determine whether the candidate meets the requirements of a true anomaly.
If the conditions check out, all samples belonging to that candidate are marked as anomalous.

At present, the \obj{analyzer} can calculate several properties of the anomaly candidate that can be used to discern its validity.
Aside from the anomaly length (in samples), amplitudes, maximum amplitude, and cumulative amplitude values are determined.
Assuming that sample $s$ belongs to category $r \in [0, m)$, and was predicted to belong to category $p \in [0,m)$, the discrepancy between mean signal values for bins $r$ and $p$ is an anomaly amplitude ($\vec{a}_s$) for that particular sample $s$ (\ref{eq:threshold:anomaly}). When amplitudes of all samples belonging to an anomaly candidate $C$ are known, the maximum amplitude (\ref{eq:threshold:max_amp}) and cumulative amplitude (\ref{eq:threshold:cum_amp}) values can be calculated.

\begin{equation}
    \vec{a}_s = \left| \frac{\vec{edges}_r + \vec{edges}_{r+1}}{2} - \frac{\vec{edges}_p + \vec{edges}_{p+1}}{2} \right|,
\label{eq:threshold:anomaly}
\end{equation}

\begin{equation}
    \mathrm{max\_amp}_C = \max_{s \in C} \vec{a}_s,
\label{eq:threshold:max_amp}
\end{equation}

\begin{equation}
    \mathrm{cum\_amp}_C = \sum_{s \in C} \vec{a}_s.
\label{eq:threshold:cum_amp}
\end{equation}

Both \obj{model} hyper-parameters and \obj{analyzer} rules are application-specific, and should be tweaked to achieve best possible performance.

\subsection{System integration}
\label{subsection:system-desc:integration}

\begin{figure}
    \centering
    \includegraphics[width=\hsize]{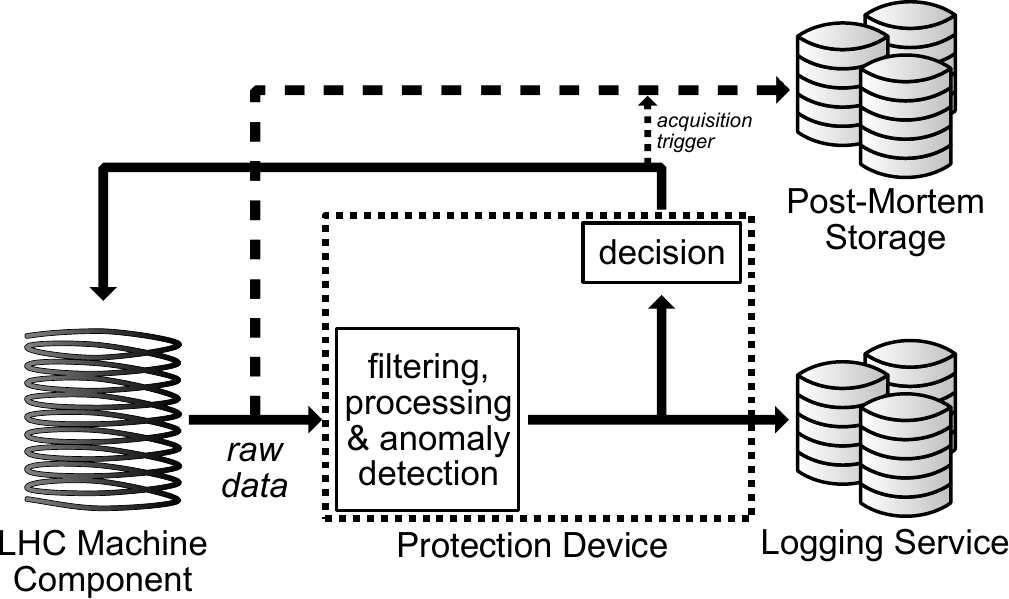}
    \caption{Currently used system. \label{fig:system-desc:current_system}}
\end{figure}

\begin{figure}
    \centering
    \includegraphics[width=\hsize]{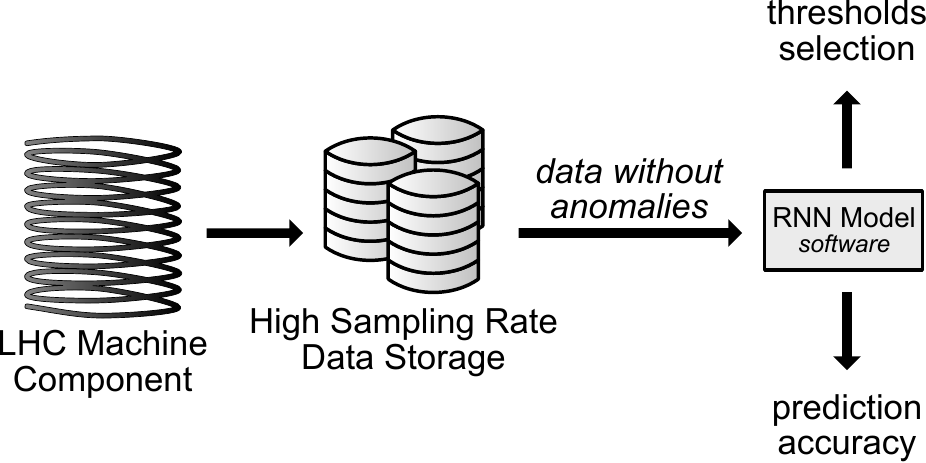}
    \caption{Data acquisition \& model training (offline). \label{fig:system-desc:offline_training}}
\end{figure}

\begin{figure}
    \centering
    \includegraphics[width=\hsize]{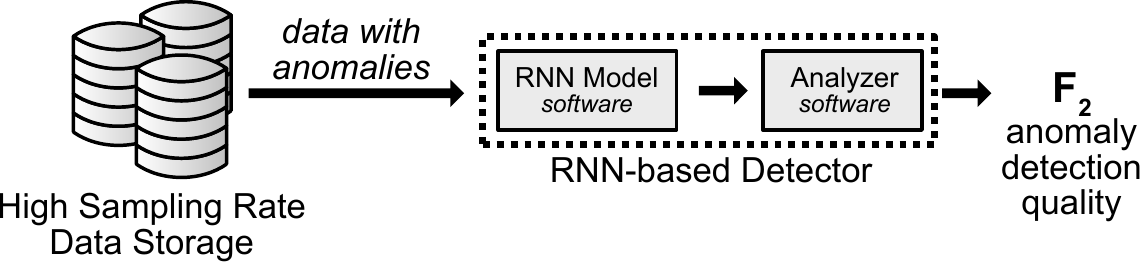}
    \caption{Model testing (offline). \label{fig:system-desc:model_testing}}
\end{figure}

\begin{figure}
    \centering
    \includegraphics[width=\hsize]{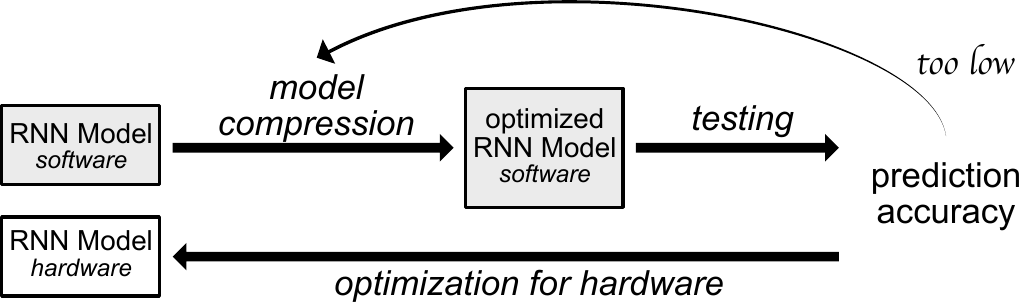}
    \caption{Design flow for future hardware implementation. \label{fig:system-desc:hardware_implementation}}
\end{figure}

\begin{figure}
    \centering
    \includegraphics[width=\hsize]{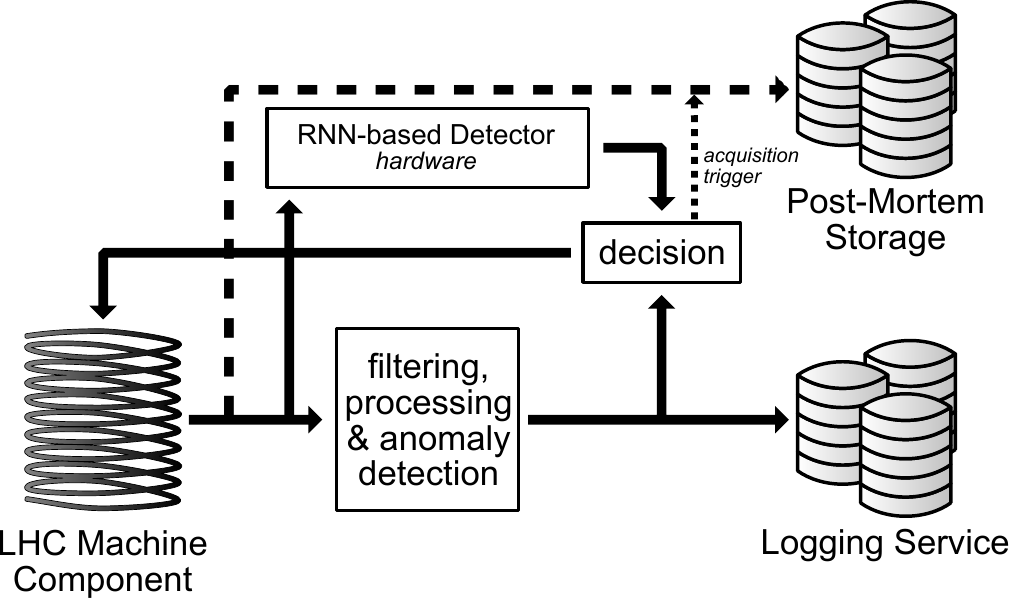}
    \caption{Proposed system. \label{fig:system-desc:proposed_system}}
\end{figure}

After the right \obj{detector} setup is chosen, it can be integrated into (in \gls{CERN} use case) magnets protection and monitoring system.
The currently used conventional system comprises, among others, hardware modules and a set of databases, as shown in Fig.~\ref{fig:system-desc:current_system}.

Voltage and current signals in powering circuits of magnets are acquired with high sampling rate (see subsection \ref{subsec:lhc:data_orig}).
Obtained raw data is then preprocessed and filtered in real-time, finally arriving in discriminating and thresholding module.
This module allows discerning whether a situation arose that needs running automatic fail-safe procedures and an expert intervention.
Described conventional solution highly depends on the expert knowledge concerning behavior and parameters of the \gls{LHC} magnets and associated equipment, which allows selecting monitoring system hyper-parameters.

Only every $n$-th sample of raw data is stored in \gls{CALS} database, where symbol $n$ denotes a decimation factor.
In actual operational scenarios there is no possibility to store all acquired data due to limited network bandwidth.
However the data collected in the \gls{CALS} database allows for later analysis and reasoning about \gls{LHC} equipment condition and behavior.
In case of trigger generation by the discriminating and thresholding module a set of samples stored in the protection device is transmitted as \glsentrylong{PM} data and is stored in dedicated database.
An original sampling rate or low decimation factor is used in this case allowing better insight into an event.

Fig.~\ref{fig:system-desc:offline_training} and \ref{fig:system-desc:model_testing} show offline \obj{model} training and testing.
The target system should use data acquired with highest possible sampling rate to ensure the required system reaction time by inferring an anomaly directly inside protection device.
For current experiments validating the approach feasibility the logging data (see subsection \ref{subsec:lhc:data_orig}) was used.
Once the \obj{model} is trained, it can be periodically updated when even more data is available.

It needs to be highlighted that every kind of magnets will need to have distinct setup.
The substantial amounts of data that could be potentially obtained should allow to very effectively train the required \obj{models}.
It is expected that a single \obj{model} instance should be sufficient for all magnets belonging to one category.
However, it will need to be verified experimentally.

The current research is focused on validation and quality evaluation of \obj{models} implemented in a high-level language.
However, once the \obj{detector} setup is determined and \obj{model} fitted, the anomaly detection algorithm will need to be implemented in hardware, for example using \gls{FPGA} platforms, to meet latency constraints (see Fig.~\ref{fig:system-desc:hardware_implementation}).

Such an implementation poses several challenges that will not be widely discussed in this paper.
However, it needs to be noted that fitting the \obj{detector} system inside an \gls{FPGA} platform that has limited computing and memory resources will require \obj{model} compression.
It also translates to the constraints on the \obj{model} size - the smaller number of \obj{model} parameters (weights), the better.
For that reason, the underlying \obj{model} hyper-parameters optimization needs to take the resources availability into account.

Fig.~\ref{fig:system-desc:proposed_system} shows the vision of a final \gls{MPS}, which includes both the proposed \gls{RNN}-based \obj{detector} and conventional solution.
Such an approach would allow increasing the reliability of the superconducting magnets monitoring system. It is also possible to use only the proposed detector module.

\section{Detector design methodology}
\label{section:methodology}

The presented \obj{detector} system has a set of hyper-parameters, that can be tweaked to achieve best results for a particular use case.
Some of them are directly influenced by the required operation macro-parameters, such as the smallest anomaly length or amplitude change, that the system should be able to detect, or the maximum response latency.

The process of tweaking the \obj{detector} setup is highly iterative, with future hardware implementation in mind.
Optimizing the \obj{model} to achieve a better accuracy usually comes hand in hand with increasing it resources consumption.
As such, contrary to the usual approach, the \obj{model} underlying the \obj{detector}, at the beginning very small, should be improved until it is just good enough for the application.

\subsection{Generic steps}
\label{subsection:methodology:steps}

\begin{algorithm*}
\caption{Methodology steps}
\label{alg:methodology:steps}
\begin{algorithmic}[2]
\Procedure{detector\_setup}{raw\_data, application\_quality\_requirements}
\State \obj{model}, \obj{analyzer}, \obj{detector}, preprocess\_config $\gets$ \Call{create}{\null}\label{line:initial:start}
\State train\_data, test\_data $\gets$ \Call{preprocess\_data}{raw\_data, preprocess\_config}
\State decimated\_data $\gets$ \Call{decimate\_data}{train\_data}
\State \method{model}{fit}{decimated\_data}
\State train\_prediction $\gets$ \method{model}{predict}{train\_data}
\State train\_anomalies $\gets$ \method{analyzer}{detect}{train\_data, train\_prediction}
\State thresholds $\gets$ \method{analyzer}{auto\_thresholds}{train\_anomalies}
\State \method{analyzer}{apply}{thresholds}
\State test\_prediction $\gets$ \method{model}{predict}{test\_data}
\State test\_anomalies $\gets$ \method{analyzer}{detect}{test\_data, test\_prediction} \label{line:initial:end}
\Statex
\While{\obj{detector}.quality < application\_quality\_requirements} \label{line:iterative:start}
    \If{\Call{is\_oversensitive}{\obj{model}}}
        \State thresholds++
    \Else
        \State best\_thresholds $\gets$ thresholds
        \State \obj{models} $\gets$ \Call{generate\_candidate\_setup}{\obj{model}, preprocess\_config} \label{line:candidates:start}
        \ForAll{\obj{candidate\_model} $\in$ \obj{models}}
            \State train\_data, test\_data $\gets$ \Call{preprocess\_data}{raw\_data, preprocess\_config}
            \State decimated\_data $\gets$ \Call{decimate\_data}{train\_data}
            \State \method{candidate\_model}{fit}{decimated\_data}
            \State train\_prediction $\gets$ \method{candidate\_model}{predict}{train\_data}
            \State train\_anomalies $\gets$ \method{analyzer}{detect}{train\_data, train\_prediction}
            \State candidate\_thresholds $\gets$ \method{analyzer}{auto\_thresholds}{train\_anomalies}
            \If{candidate\_thresholds < best\_thresholds}
                \State \obj{best\_candidate} $\gets$ \obj{candidate\_model}
                \State best\_thresholds $\gets$ candidate\_thresholds
            \EndIf
        \EndFor
        \State \obj{model} $\gets$ \obj{best\_candidate}  \label{line:candidates:end}
        \State \method{analyzer}{apply}{best\_thresholds}
        \State test\_prediction $\gets$ \method{model}{predict}{test\_data}
        \State test\_anomalies $\gets$ \method{analyzer}{detect}{test\_data, test\_prediction}
    \EndIf
\EndWhile \label{line:iterative:end}
\Statex
\State \Return \obj{detector}
\EndProcedure
\end{algorithmic}
\end{algorithm*}

In the initial phase (\alglines{alg:methodology:steps}{line:initial:start}{line:initial:end}), the data is preprocessed (normalized, quantized, formatted according to \obj{model} needs and split into training and testing sets) and, if possible and to reduce computation time, decimated.

Next, the starting \obj{model} is fitted with decimated data and then used to make predictions on the training set.
Predictions obtained from the \obj{model} are then used by the \obj{analyzer} to detect anomalies.

Since anomalies are detected in training set (where target system should report none), they can be used to adjust \obj{analyzer} thresholds automatically.
Procedure for automatic thresholds and rules selection is described in the following subsection.
Finally, the \obj{model} is used to make predictions on testing data, which are then used for anomaly detection.

The iterative phase (\alglines{alg:methodology:steps}{line:iterative:start}{line:iterative:end}) starts with \obj{detector} quality evaluation (using Precision, Recall and F-score metrics, see subsection \ref{subsection:experiments:quality-measures}). 
Exact quality evaluation depends on application needs, e.g., there can be applications where a lower number of false positives is more important than a lower number of false negatives.

If an amount of false positives is high and real anomalies are (using measures like a length or a cumulative amplitude) bigger than those incorrectly reported anomalies, the model can be considered as oversensitive.
This situation can be addressed by increasing \obj{analyzer} threshold values, especially those where the gap between real true and false anomalies is significant.

If thresholds adjustment is impossible (e.g., true anomalies measurements are similar to those of false anomalies, or the \obj{model} is not accurate enough, resulting in high false negatives number), the only way to improve the \obj{detector} quality is by changing the preprocessing or improving the underlying \obj{model}.

After setup is improved, the \obj{detector} quality evaluation can begin anew.

\subsection{Automatic rules adjustment}
\label{subsection:methodology:thresholds}

\begin{figure*}
    \centering
    \includegraphics[width=0.75\hsize]{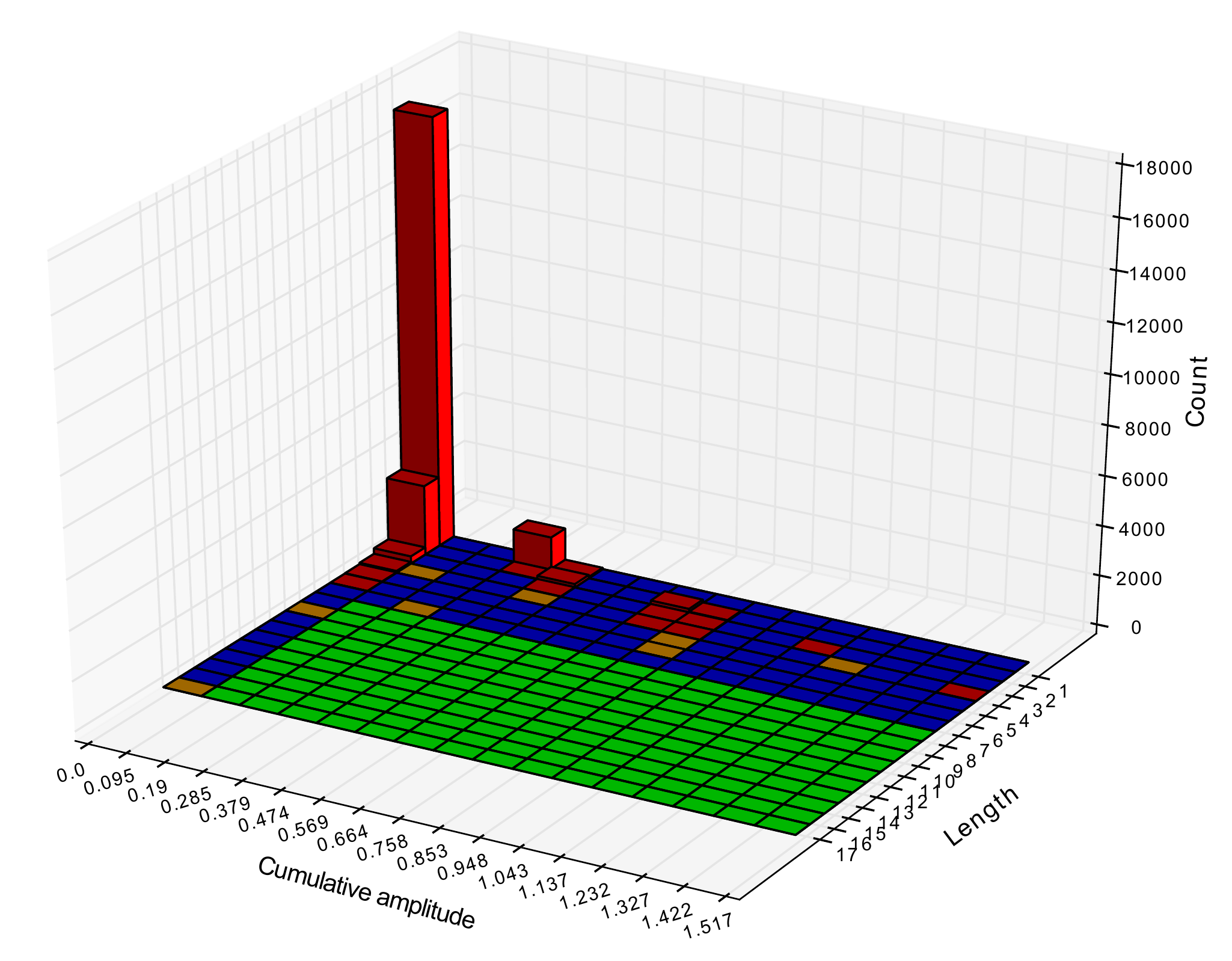}
    \caption{Example chart showing false anomalies in first 200k training samples; \textit{saved area} marked in green. \label{fig:collimation_chart}}
\end{figure*}

\begin{algorithm*}
\caption{Automatic thresholds and rules selection}
\label{alg:methodology:auto_thresholds}
\begin{algorithmic}[2]
\Procedure{\obj{analyzer}.auto\_thresholds}{train\_anomalies} \label{line:auto_thresholds:start}
    \State possible\_thresholds $\gets$ \Call{supported\_thresholds}{\null}
    \ForAll{threshold $\in$ possible\_thresholds}
        \State maxima[threshold] $\gets$ \Call{max}{train\_anomalies[threshold]}
        \State best\_combination[threshold] $\gets$ 0
        \State bins[threshold] $\gets$ \Call{compartmentalize}{train\_anomalies[threshold]}
    \EndFor
    \State best\_combination[possible\_thresholds[0]] $\gets$ maxima[possible\_thresholds[0]]
    \State best\_area $\gets$ 0 \label{line:auto_thresholds:setup}
    \State possible\_combinations $\gets$ \Call{combine}{bins, possible\_thresholds} \label{line:auto_thresholds:combinations}
    \ForAll{combination $\in$ possible\_combinations}
        \State is\_valid\_combination $\gets$ \True        
        \ForAll{anomaly $\in$ train\_anomalies}
            \If{\Call{above\_thresholds}{anomaly, combination}}
                \State is\_valid\_combination $\gets$ \False
                \State \Break
            \EndIf
        \EndFor
        \If{is\_valid\_combination}
            \State area $\gets$ \Call{calculate\_saved\_area}{combination, maxima}
            \If{area > best\_area}
                \State best\_area $\gets$ area
                \State best\_combination $\gets$ combination
            \EndIf
        \EndIf
    \EndFor \label{line:auto_thresholds:end}
    \State rules $\gets$ []
    \ForAll{threshold $\in$ possible\_thresholds}
        \If{best\_combination[threshold] > 0}
            \State rules.\Call{push}{threshold}
        \EndIf
    \EndFor
    \State \Return best\_combination, rules
\EndProcedure
\end{algorithmic}
\end{algorithm*}

The proposed automatic analyzer rules selection procedure, described in Alg.~\ref{alg:methodology:auto_thresholds}, is conceptually simple.
Its objective is to find such a combination of threshold values that will guarantee filtering out all false anomalies found in training set.
At the same time, it should not be too greedy, in case true anomalies are close to false ones.

For example, most of the false anomalies can be shorter than a certain length, while at the same time spanning full cumulative amplitude range.
Simultaneously, there can exist a small number of long false anomalies which are low in cumulative amplitude.
It can be then surmised that any longer (but still within range) anomaly with higher (in range) cumulative amplitude would be a true one and should not be filtered out.
If only greedy, essential thresholds were applied (,,anomaly is true if its length or cumulative amplitude or amplitude is bigger than the relevant maximum found for false anomalies''), the anomaly as mentioned earlier would not match any of those criteria and therefore would not be detected.

In Fig.~\ref{fig:collimation_chart}, the example visualization of false anomaly properties is shown.
It is a 2-D histogram, with an amount of bins equal to the amount of possible discrete values (for small anomaly length ranges) or calculated using numpy 'sturges' algorithm \citep{online:numpy:sturges}.
Assuming there are only two thresholds possible (length and cumulative amplitude, for example), the algorithm task is to find such values, that will result in the largest possible \textit{saved area} (marked in green in Fig.~\ref{fig:collimation_chart}).
The \textit{saved area} represents an additional space (aside from areas beyond the found maximums) in which anomaly detection will be possible.
This reasoning is then generalized to an arbitrary number of parameters.

Initially, the algorithm sets the best threshold combination to contain a value only for the first threshold, with others set to \num{0}.
The first threshold value is based on the maximum value appearing in anomalies, and the \textit{saved area} is set to \num{0} (\alglines{alg:methodology:auto_thresholds}{line:auto_thresholds:start}{line:auto_thresholds:setup}).

In the next part, possible threshold combinations are checked (\alglines{alg:methodology:auto_thresholds}{line:auto_thresholds:combinations}{line:auto_thresholds:end}).
For every detected anomaly, it is checked if a particular threshold combination can be used to filter it out.
If the combination can be used to filter all anomalies, it is considered valid and \textit{saved area} for that particular combination is calculated.
The best threshold combination is the one with the highest \textit{saved area}.

\subsection{Setup improvement}
\label{subsection:methodology:setup}

A crucial part of improving detector setup is selecting the optimum underlying model (\alglines{alg:methodology:steps}{line:candidates:start}{line:candidates:end}).
The number of \gls{NN} models hyper-parameters optimization methods is rapidly growing, starting with the heuristic-based approaches and moving toward ones utilizing \gls{RL} and Bayesian algorithms \citep{zoph2017neural, bello2017neural, brochu2010tutorial, li2016learning}.
The description of those approaches is beyond the scope of this paper, but it needs to be pointed out that most of those algorithms are not created with hardware implementation in mind.
The authors are researching an automatic, resource-aware \gls{NN} models hyper-parameters optimization, the preliminary concept of which is described in \citep{wielgosz2016observer}, to address this issue.
\section{Experiments}
\label{section:experiments}

The experiments with both evenly-spaced grid and adaptive one were conducted and compared with the standard \gls{SVM}-based solution.
Collected results allowed to judge the effectiveness and efficiency of the proposed solution.

\subsection{Dataset}
\label{subsection:experiments:dataset}

\begin{table}
  \centering
  \caption{Properties of the used dataset. \label{tab:experiments:data}}
  \begin{tabular}{lccc}
    \toprule
    & \multicolumn{3}{c}{samples (in millions)} \\
    \cmidrule{2-4}
    series & training & testing & total\\
    \midrule
    h1144 & \num{\sim 3.8} & \num{\sim 1.2} & \num{\sim 5} \\
    h1011 & \num{\sim 5.1} & \num{\sim 0.4} & \num{\sim 5.5} \\
    h1451 & \num{\sim 4.5} & \num{\sim 0.5}  & \num{\sim 5} \\
	h1819 & \num{\sim 5.7} & \num{\sim 0.3}  & \num{\sim 6} \\ 
    \bottomrule
  \end{tabular}
\end{table}

The data used in the experiments were acquired during \gls{HL} magnets training. However, the developed methods will also be applicable during the normal machine operation (during the \gls{LHC} cycle, when the magnet works at the nominal current).
Those magnets are still in a testing and training phase, with their characteristics being checked and operation parameters verified.

The magnet training consists of repeated magnet runs during which the current is slightly increased until they quench under control, with the aim of stabilizing the magnet at its design specification (ultimate current).
The data was obtained from the short model of the \SI{11}{\tesla} dipole magnet (MBHSP105) with a single aperture (see subsection~\ref{subsec:lhc:data_orig} for detailed description).
Each of the series contains four data channels, with first two representing the voltages on magnet coils, third - the current measurement and fourth - the compensated signal (sum of the first two) (Fig.~\ref{fig:lhc:meas_setup}).

To obtain actual voltage values from the signals, they need to be multiplied by gain of analogue stage $G~=~\SI[per-mode=symbol]{5}{\volt\per\volt}$ and conversion factor of \gls{ADC} $\mathit{LSB}~=~\SI[per-mode=symbol]{9.5348}{\micro\volt\per\bit}$.
The current signal is acquired from the voltage output of the \gls{DCCT} installed on the power converter, and the related value is obtained multiplying the voltage signal from the \gls{DCCT} by a conversion factor of \SI[per-mode=symbol]{2}{\kilo\ampere\per\volt}.
For the experiments, only the first three channels were used.

The collected data is divided into four series (h1011, h1144, h1451 and h1819), all coming from the same magnet during the different training runs.
Each of the series contains an extended period of the normal operation (ramp up of the magnet), followed by an anomaly (quench) and results of a power abort procedure.

Each of the series was then split into two parts, one containing only normal operation data (training set) and the second one containing the anomaly and power abort in addition to normal operation (testing set; see Tab.~\ref{tab:experiments:data} for details).

Both the quenches and the power abort fragments were annotated as anomalies when measuring the detector performance since both contain phenomena that the model has never seen.

Since there are only a few real anomalies available, to further examine the detector performance, the tests sets containing synthetic anomalies were created.
For that purpose, the normal operation part of h1011 series was augmented with a thousand of synthetic anomalies added.

In the synthetic set I, introduced anomaly is a unit step impulse with the duration of \num{100} samples (see Fig.~\ref{fig:experiments:false_negative} and \ref{fig:experiments:false_positive}).
The synthetic set II is similar, only with unit step impulse duration set to \num{50} samples.

\subsection{Preprocessing}
\label{subsection:experiments:preprocessing}

The acquired data needs to be initially prepared to be used for recurrent neural network training.
The first step is the signals normalization to $[0, 1]$ range, using all of the available data.
The normalized data is saved to be reused in all the experiments.
Normal operation data is used for model training and validation, while data containing anomalies and power abort is used for complete detector setup testing.

In the next step, based on \var{in\_grid}, \var{out\_grid}, \var{in\_algorithm} and \var{out\_algorithm} configuration values, the grid edges are calculated.
They will later be used to quantize input and output signals.

Following that the data structuring is done.
For all the data points that have the required history length (the sample index in series is greater than or equal to \var{look\_back} + \var{look\_ahead}) tensors containing that history (with \var{look\_back} length) for all three used channels are created.
At the same time, this historical data is quantized, using previously calculated edges.
It needs to be highlighted that, unlike when working with statistical models such as \gls{SVM}, the data used for training is overlapping.
Simultaneously, the output data tensor using 'one hot' encoding is created, with length equal to the \var{out\_grid} parameter.
Voltage 0 signal was selected as the prediction target.

After history tensors and linked output categories are prepared for each data series, they are all mixed.
For the actual experiment, a fraction of data specified by the $\mathtt{samples\_percentage} \in [0, 1]$ is randomly chosen.

\subsection{Quality measures}
\label{subsection:experiments:quality-measures}

Several standard quality measures were used to compare models and detector setups, as well as compare proposed solution performance with alternative approaches.

\subsubsection{Model quality}
\label{subsection:experiments:model-quality}

The metric used for measuring the underlying \gls{GRU} model is accuracy.
Given the values $t$ and $f$ representing, respectively, an amount of correctly and incorrectly classified samples, the accuracy can be defined as in (\ref{eq:accuracy}):

\begin{equation}
\mathrm{accuracy} = \frac{\mathit{t}}{\mathit{t} + \mathit{f}}.
\label{eq:accuracy}
\end{equation}

\subsubsection{Detector quality}
\label{subsection:experiments:detector-quality}

A switch from measuring the quality on a per-sample basis to a per-anomaly basis is needed to score the detector performance.
This need is especially true considering the rarity of anomalies, where the number of samples belonging to the 'anomaly' category is insignificant when compared with the number of normal operation samples.
In such a case, any metric incorporating total number of samples (like accuracy) would not provide any meaningful information about anomaly detection capabilities.

Scoring the performance on a per-anomaly basis, on the other hand, needs some well-defined rules for the metrics to be useful.
The most straightforward question that needs to be answered is ,,when the detected anomaly (positive) is considered true?''.
In this paper, a detected anomaly is considered to be true positive if any part of it overlaps with the real anomaly.
What follows, if several detected anomalies are overlapping a single real one, all of them are considered true.
This behavior also occurs in reverse, if a single detected anomaly spans several real ones, all of them are considered to be found.

Depending on the application needs, it may be crucial to be able to qualify the detection quality further.
An attempt to develop the more comprehensive anomaly detection metrics can be found for example in \citep{lavin2015evaluating}.

It needs to be noted that, due to continuous nature of detector operation, it is impossible to define a true negative.
Such a notion would not only require artificial splitting of time series into windows of arbitrary length and overlap but also contradicted the purpose of the switch from per-sample based metrics to per-anomaly ones.
This lack of true negatives narrows down the available standard quality measures.

The selected quality metrics should reflect the application needs.
In case of the \gls{HL} data, it is crucial to find all anomalies, since undetected faults may lead to huge disaster in the LHC tunel and in the consequence costly repairs and long accelerator shutdowns.
On the other hand, false positives reflect the machine availabilty which is a crucial operational parameter of the accelerator.
The two metrics, measuring those features, are recall (\ref{eq:recall}), also called sensitivity, and precision (\ref{eq:precision}), respectively:
\begin{equation}
    \mathrm{recall} = \frac{\mathit{tp}}{\mathit{tp} + \mathit{fn}},
    \label{eq:recall}
\end{equation}
\begin{equation}
    \mathrm{precision} = \frac{\mathit{tp}}{\mathit{tp} + \mathit{fp}},
    \label{eq:precision}
\end{equation}
where:
\begin{itemize}
    \item $\mathit{tp}$ -- true positive -- item correctly classified as an anomaly,
    \item $\mathit{fp}$ -- false positive -- item incorrectly classified as an anomaly,
    \item $\mathit{fn}$ -- false negative -- item incorrectly classified as a part of normal operation.
\end{itemize} 

To combine those two metrics into a single one that can be directly applied for anomaly detection solutions comparison, an F-measure is used (\ref{eq:f_measure}). The $\beta$ parameter controls the recall importance in relevance to the precision:

\begin{equation}
    \mathrm{F}_\beta = (1 + \beta^2) \cdot \frac{\mathrm{recall} \cdot \mathrm{precision}}{\mathrm{recall} + \beta^2 \cdot \mathrm{precision}}.
    \label{eq:f_measure}
\end{equation}

During the detector performance experiments two $\beta$ values were used, \num{1} and \num{2}, to show the impact of the recall on the final score.

\subsection{Methodology validation}
\label{subsection:experiments:methodology}

\begin{figure}
    \centering
    \includegraphics[width=\hsize]{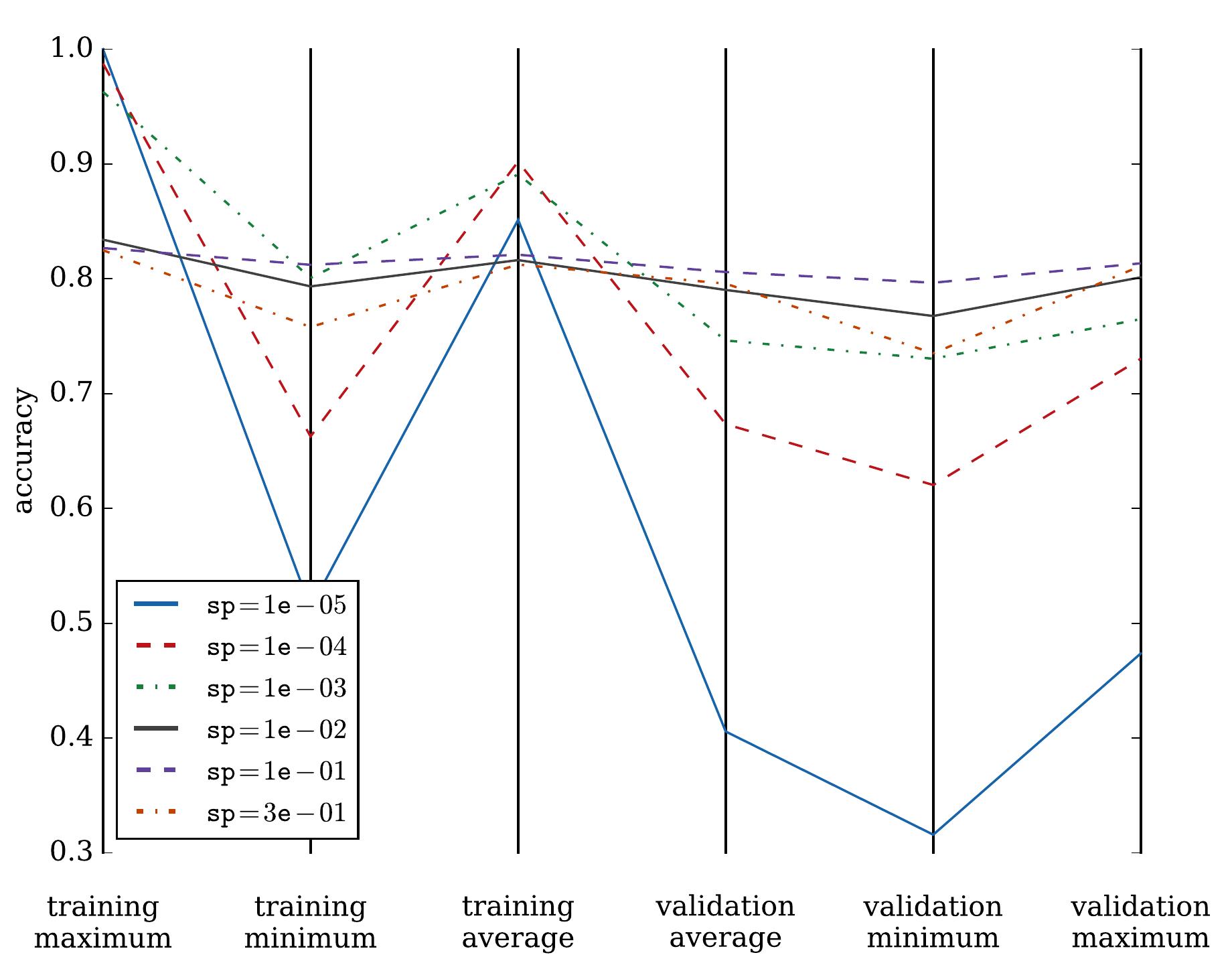}
    \caption{Influence of \var{samples\_percentage} (\var{sp}) on model accuracy (\var{look\_back}{=10}, \var{in\_grid}{=10}, \var{out\_grid}{=10}, average over various \var{cells} values). \label{fig:experiments:sp_stats}}
\end{figure}

\begin{figure}
    \centering
    \includegraphics[width=\hsize]{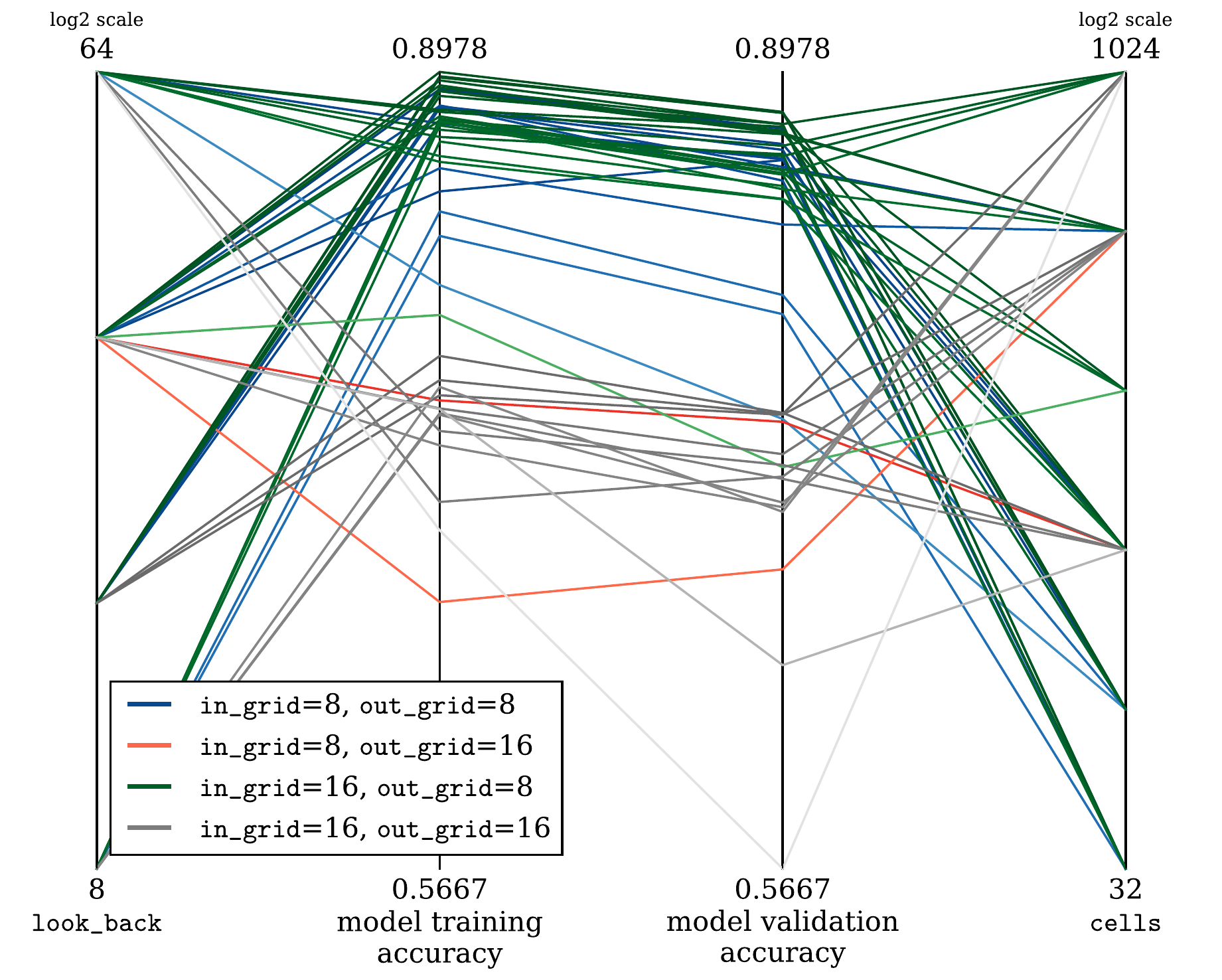}
    \caption{Influence of \var{look\_back} and \var{cells} on model accuracy for various \var{in\_grid} and \var{out\_grid} combinations (\var{samples\_percentage}{=0.01}). Darker colors indicate higher model validation accuracy. \label{fig:experiments:hp_trends}}
\end{figure}

\begin{figure*}
    \centering
    \includegraphics[width=\hsize]{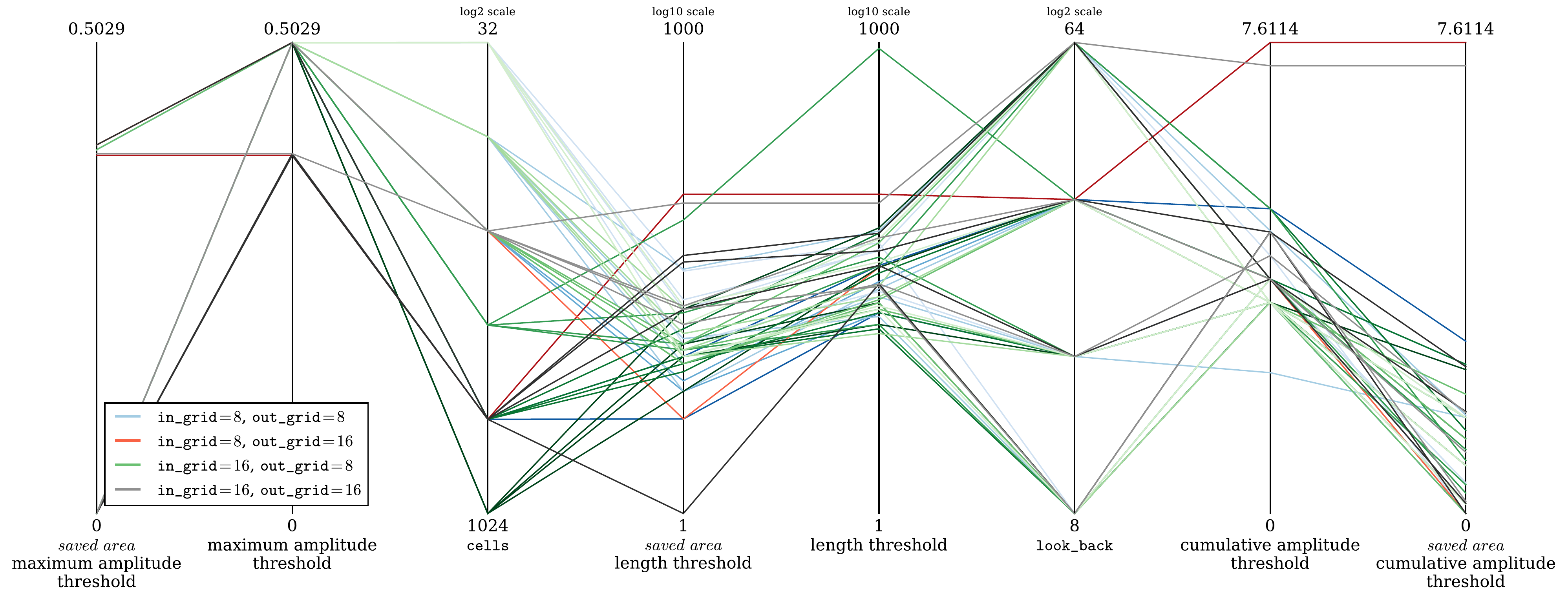}
    \caption{Influence of \var{look\_back} and \var{cells} on automatically calculated thresholds (\var{samples\_percentage}{=0.01}). Darker colors indicate bigger model (bigger \var{cells} values). See Tab.~\ref{tab:experiments:thresholds_stats} for statistics. \label{fig:experiments:thresholds}}
\end{figure*}

\begin{table}
    \centering
    \caption{Automatically calculated thresholds statistics. \label{tab:experiments:thresholds_stats}}
    \begin{tabular}{lcccc}
    \toprule
    threshold & min & max & median & mean \\
    \midrule
    length & 14 & 918 & 29.5 & 56 \\
    \textit{saved~area} length & 1 & 108 & 12 & 19.76 \\
    cumulative amp. & 2.28 & 7.61 & 3.79 & 3.92 \\
    \makecell[l]{\textit{saved~area}\\cumulative amp.} & 0 & 7.61 & 1.01 & 1.3 \\
    maximum amplitude & 0.38 & 0.5 & 0.5 & 0.49 \\
    \makecell[l]{\textit{saved~area}\\maximum amplitude} & 0 & 0.39 & 0 & 0.07 \\
    \bottomrule
    \end{tabular}
\end{table}


\begin{figure}
    \centering
    \includegraphics[width=\hsize]{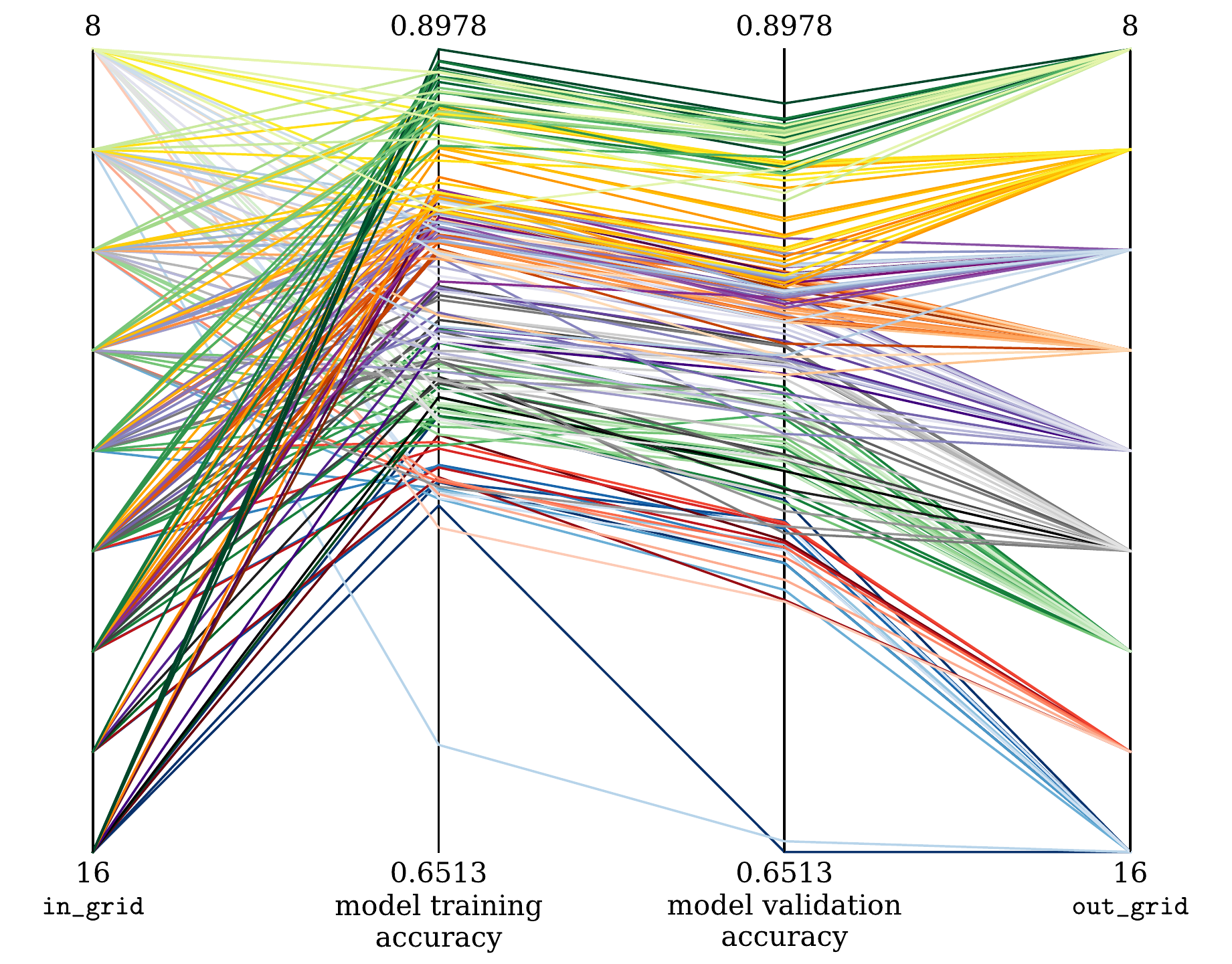}
    \caption{Influence of grid sizes on model accuracy (\var{look\_back}{=32}, \var{samples\_percentage}{=0.01}). Darker color shades indicate bigger \var{in\_grid}, while colors themselves signify \var{out\_grid} values. \label{fig:experiments:in_out_grid}}
\end{figure}

\begin{figure}
    \centering
    \includegraphics[width=\hsize]{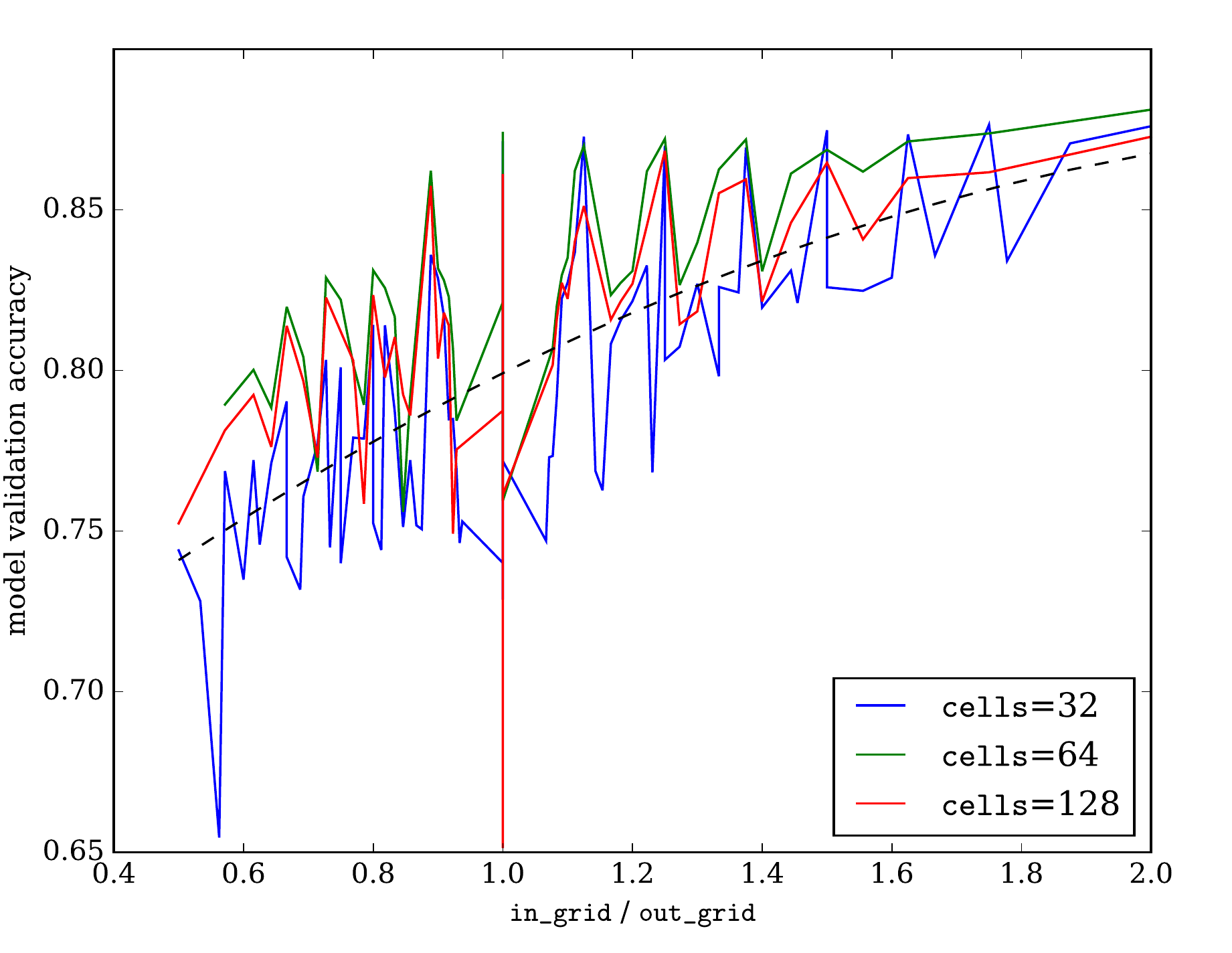}
    \caption{Influence of grid sizes ratio on model accuracy (\var{look\_back}{=32}, \var{samples\_percentage}{=0.01}). \label{fig:experiments:in_out_grid_ratio}}
\end{figure}

Experiments involving neural networks are usually very resource-consuming.
To reduce computational cost, it may be beneficial to at first train model on (small) representative fraction of available training data.
A random sweep with increasing percentage values was conducted to select such a percentage used in later experiments.
The sweep results can be seen in Fig.~\ref{fig:experiments:sp_stats}.
The data fraction can be considered to be big enough when, for a given model, training and validation accuracy is similar.
Based on the experiment results, it can be seen that the above condition is true starting with \SI{1}{\percent} of the original dataset size (\var{samples\_percentage}{=0.01}) -- in the visualization, the line connecting those two accuracy values is nearly horizontal.
Moreover, the relative difference between average accuracy achieved for higher \var{samples\_percentage} values is minimal.
Therefore, the \SI{1}{\percent} value was determined to be sufficient for the following tests.

Fig.~\ref{fig:experiments:hp_trends} shows the relationship between four  hyper-parameters: history window length (\var{look\_back}), number of \gls{GRU} model cells and \var{in\_grid}/\var{out\_grid} values.
It can be observed that model performance mainly depends on grid sizes (with best results achieved for \var{in\_grid}{=16} and \var{out\_grid}{=8}), with \var{look\_back} and model size (\var{cells}) values having a surprisingly small impact.
It is, however, worth noting that smaller models with smaller \var{look\_back} values tend to have better performance than those with one of the parameters closer to the upper tested limit.

Fig.~\ref{fig:experiments:thresholds} visualizes relationships between \var{look\_back}, model validation accuracy, and calculated threshold parameters.
The grids sizes influence on length and cumulative amplitude thresholds is small, but noticeable, especially affecting maximum false anomaly length, which in turns affects \textit{saved area} length threshold.
The \var{look\_back} values seem to play a significant role in determining model capabilities -- smaller values tend to result in lower maximum false anomalies length.
It can also be seen that maximum false anomaly amplitude (topping up around \num{0.5}) depends almost entirely on \var{out\_grid}.

As an additional experiment, authors conducted a more thorough research into the impact of grid sizes on model accuracy (Fig.~\ref{fig:experiments:in_out_grid} and \ref{fig:experiments:in_out_grid_ratio}).
It turns out that \var{in\_grid}/\var{out\_grid} ratio is very visibly related to model accuracy, with lower ratio values lowering the model performance. 
However, if the higher ratio cannot be achieved, it is better to use lower \var{out\_grid} value.

\subsection{Detector performance}
\label{subsection:experiments:detector}

\begin{table*}
    \caption{Detector performance results. \label{tab:experiments:detector_performance}}
    \begin{tabular}{llccccc}
         \toprule
         \multicolumn{2}{c}{} & \multicolumn{5}{c}{Setup} \\
         \cmidrule{3-7}
         & & \var{best\_length} & \var{best\_cum\_amp} & \var{best\_max\_amp} & \var{best\_accuracy} & \var{balanced} \\
         \midrule
         \multirow{4}{4.4em}{Parameters} & \var{in\_grid} & 16 & 8 & 16 & 16 & 16 \\
         & \var{out\_grid} & 8 & 8 & 16 & 8 & 8 \\
         & \var{look\_back} & 16 & 16 & 16 & 32 & 8 \\
         & \var{cells} & 64 & 64 & 512 & 64 & 64 \\
         \midrule
         \multirow{2}{4.4em}{Model accuracy} & train & 0.8955 & 0.8828 & 0.7634 & \textbf{0.8978} & 0.8756  \\
         & validation & 0.8809 & 0.8654 & 0.7553 & \textbf{0.8812} & 0.8551 \\
         \midrule
         \multirow{3}{4.4em}{False anomalies maximums} & \var{length} & \textbf{14} & 24 & 38 & 24 & 29 \\
         & \var{cum\_amp} & 3.4104 & \textbf{2.2761} & 3.7890 & 3.4110 & 3.4110 \\
         & \var{max\_amp} & 0.5026 & 0.5026 & \textbf{0.3824} & 0.5026 & 0.5026 \\
         \midrule
         \multirow{3}{4.4em}{\textit{Saved area} thresholds} & \var{length} & 10 & \textbf{9} & 20 & 14 & 11 \\
         & \var{cum\_amp} & 1.5503 & 1.5520 & 1.6474 & \textbf{0.7755} & \textbf{0.7755} \\
         & \var{max\_amp} & 0 & 0 & 0 & 0 & 0 \\
         \midrule
         \multirow{4}{4.4em}{Real set} & recall & 1.0 & 1.0 & 1.0 & 1.0 & 1.0 \\
         & precision & \textbf{1.0} & 0.0046 & 0.0005 & \textbf{1.0} & \textbf{1.0} \\
         & $\mathrm{F}_1$ & \textbf{1.0} & 0.0091 & 0.0009 & \textbf{1.0} & \textbf{1.0} \\
         & $\mathrm{F}_2$ & \textbf{1.0} & 0.0225 & 0.0023 & \textbf{1.0} & \textbf{1.0} \\
         \midrule
         \multirow{4}{4.4em}{Synthetic set~I} & recall & 0.9974 & 0.9950 & 0.6848 & \textbf{0.9979} & 0.8312 \\
         & precision & \textbf{1.0} & \textbf{1.0} & \textbf{1.0} & 0.9986 & \textbf{1.0} \\
         & $\mathrm{F}_1$ & \textbf{0.9987} & 0.9975 & 0.8129 & 0.9982 & 0.9078 \\
         & $\mathrm{F}_2$ & 0.9979 & 0.9960 & 0.7309 & \textbf{0.9980} & 0.8602 \\
         \midrule
         \multirow{4}{4.4em}{Synthetic set~II} & recall & \textbf{0.8685} & 0.8184 & 0.4340 & 0.8643 & 0.5045 \\
         & precision & \textbf{1.0} & 0.9988 & \textbf{1.0} & 0.9977 & \textbf{1.0} \\
         & $\mathrm{F}_1$ & \textbf{0.9296} & 0.8996 & 0.6053 & 0.9262 & 0.6706 \\
         & $\mathrm{F}_2$ & \textbf{0.8920} & 0.8490 & 0.4894 & 0.8880 & 0.5600 \\
         \bottomrule
    \end{tabular}
\end{table*}

\begin{figure*}
    \centering
    \includegraphics[width=0.8\hsize]{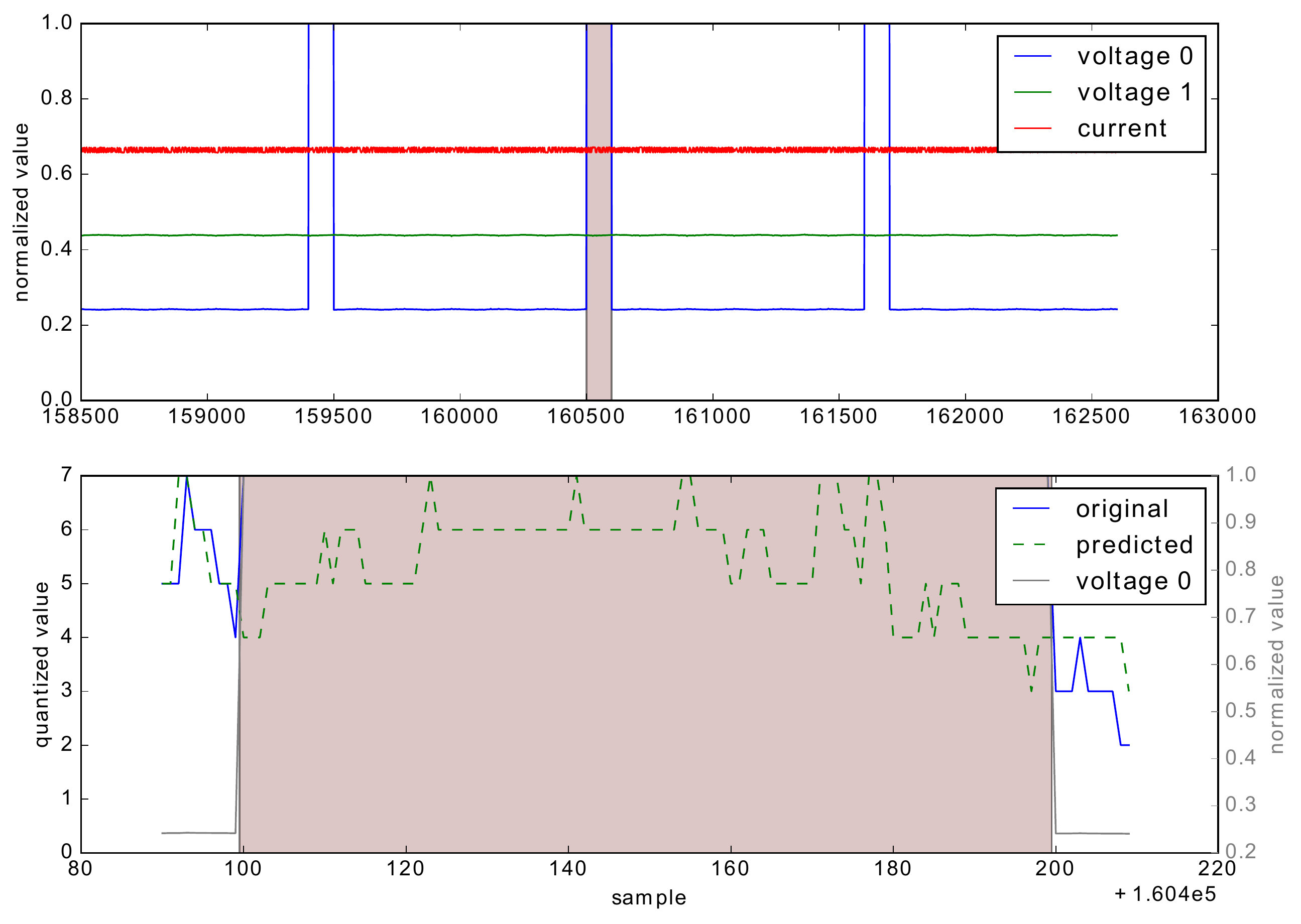}
    \caption{False negative (samples 160500 -- 160600) anomaly in synthetic set I missed by \var{best\_accuracy} detector setup. \label{fig:experiments:false_negative}}
\end{figure*}

\begin{figure*}
    \centering
    \includegraphics[width=0.8\hsize]{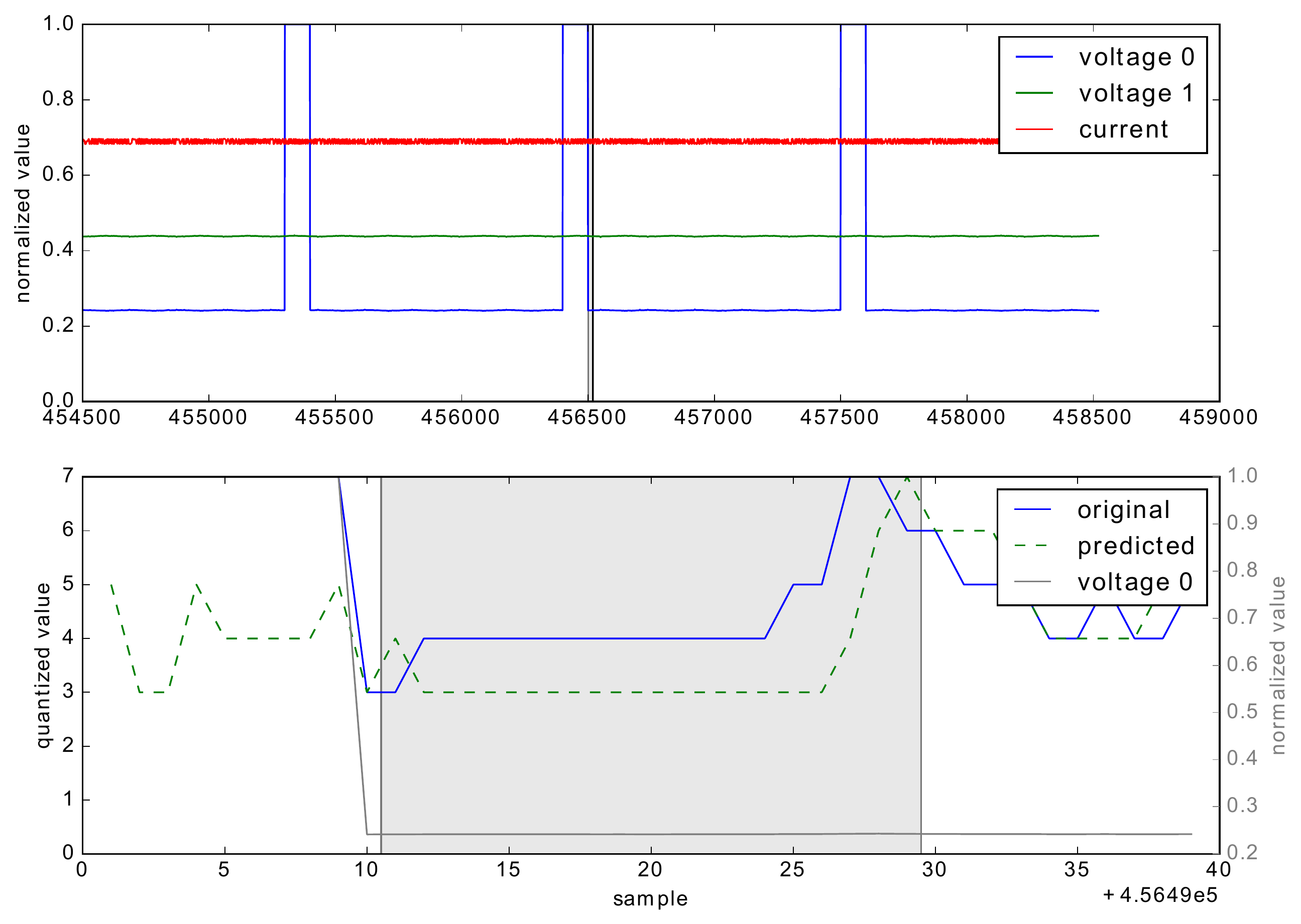}
    \caption{False positive (samples 456501 -- 456520) anomaly in synthetic set I found by \var{best\_accuracy} detector setup. \label{fig:experiments:false_positive}}
\end{figure*}

Five of the setups were selected using different criteria to test detector performance:
\begin{itemize}
    \item \var{best\_length} -- setup with the lowest maximum false anomaly length,
    \item \var{best\_cum\_amp} -- the lowest maximum false anomaly cumulative amplitude,
    \item \var{best\_max\_amp} -- the lowest maximum false anomaly amplitude,
    \item \var{best\_accuracy} -- the highest model validation accuracy
    \item \var{balanced} -- setup with relatively good accuracy, low maximum values of length and cumulative amplitude, as well as \textit{saved area} length and cumulative amplitude thresholds.
\end{itemize}
Exact setup parameters, as well as performance results, are described in Tab.~\ref{tab:experiments:detector_performance}.

Looking only at the results on the real data test set, it may seem that \var{best\_length}, \var{best\_accuracy}, and \var{balanced} setups perform equally well.
The performance of the \var{best\_cum\_amp} and \var{best\_max\_amp} ones is abysmal, which outright disqualifies them from being used with this particular data.

Looking at the results for test sets containing unit step impulses, however, brings out a bit different picture.
The \var{balanced} setup is not performing nearly as well, with a significant drop in the recall value even for long impulses and meager score for shorter ones.
The \var{best\_length} setup scores are still very high, with \var{best\_accuracy} ones slightly outperforming it regarding recall and $\mathrm{F}_2$ for longer impulses.

Fig.~\ref{fig:experiments:false_negative} shows the example of false negative, missed by the \var{best\_accuracy} setup in synthetic set I.
When predicted and real quantized anomaly values in the selected section of the signal are compared, it can be seen that in several points the predicted values fall into the highest range.
From the detector point of view it means that the model, after some small error, correctly predicted current value, and the anomaly candidate can be disregarded.
Such a situation occurs as a result of the discriminating thresholds being too high in this particular case.

One of the ways in which this problem possibly may be mitigated is an \var{out\_grid} increase.
Direct application of this solution, however, severely affects the model accuracy (as shown in Fig.~\ref{fig:experiments:in_out_grid}) --  \var{out\_grid} increase needs to come in hand with other parameters, especially \var{in\_grid}, adjustment.

Another way involves changing the way analyzer confirms or rejects anomaly candidates.
As mentioned, currently candidate is rejected whenever a true prediction is made, unless the configured thresholds were passed first.
An alternative approach, subject to future research, could involve tracking a ratio of true vs. false predictions or introducing required true predictions threshold for candidate rejection.

Fig.~\ref{fig:experiments:false_positive} shows the example false positive, found in the synthetic set I using the \var{best\_accuracy} setup.
It can be seen that the false positive was reported soon after the synthetic anomaly occurred, so it can be assumed that incoming anomalous signal affected the model predictions.
In the real-world scenario, this should not be a problem, since first detected anomaly would probably trigger a failsafe mechanism.
In cases where anomalies should be detected even if they occur one after another, a solution involving a small ignored window, equal in length to a half or a whole \var{look\_back} value, could probably be implemented.
How such a mechanism would affect the whole detector performance needs to be researched.

Overall, it seems that for \gls{HL} data analysis, the setups selected based on the lowest maximum false anomaly length may yield the best performance, with the ones based on best accuracy being nearly as good.

\subsection{Alternative approach: OC-SVM}
\label{subsection:experiments:svm}

\begin{figure*}
    \centering
    \includegraphics[width=0.7\textwidth]{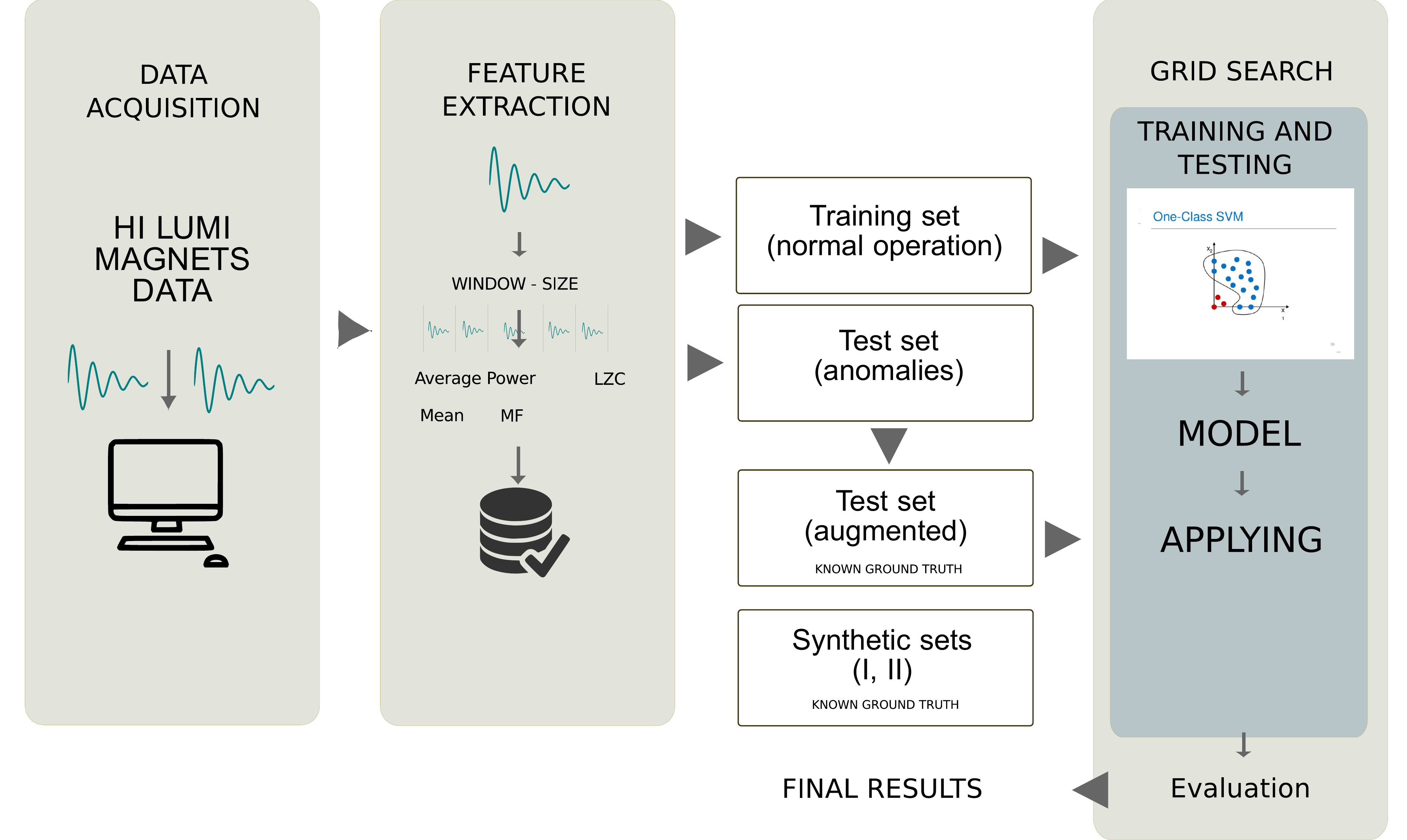}
    \caption{Overall block diagram summarizing the main processing stages and anomaly detector based on \gls{OC-SVM}. \label{fig:experiments:oc_svm_diagram}}
\end{figure*}

Several implementations of anomaly detection systems based on \gls{OC-SVM} were proposed with promising results \cite{Ma2003Time, Zhang2007One, Su2015Anomaly, s141120713}.
These algorithms are trained offline to subsequently work online.
In this subsection, the comparison of \gls{OC-SVM} models with the proposed \gls{GRU}-based system is presented.

Some properties of the experimental setup needed to be changed as required by the nature of the \gls{OC-SVM}.
Therefore, the \gls{HL} data was preprocessed accordingly.
In following paragraphs the \gls{OC-SVM} algorithm, the preprocessing and the experimental setup with results are described in the details.

\subsubsection{\glsentrylong{OC-SVM}}

\glspl{OC-SVM} are a particular case of \gls{SVM}, that can be trained with unlabeled data.
Therefore, they are an example of unsupervised machine learning techniques.
In \gls{OC-SVM} the support vector model is trained on data that has only one class, the ,,normal'' class, which infers the properties of normal cases.
As such, after training, the examples ,,unlike normal examples'' can be detected.
In an anomaly detection field, this is especially useful, as there are many different situations where the training examples are scarce (such as fraud detection, or network intrusion).

In the experiments, the \gls{OC-SVM} by Sch\"{o}lkopf et al. \citep{scholkopf2000support, Scholkopf:2001:ESH:1119748.1119749} implemented using Sklearn Python library was used.
This \gls{SVM} separates all the data points from the origin (in feature space $F$) and maximizes the distance from this hyperplane to the origin.
This operation results in a binary function which captures regions in the input space where the probability density of the data is positioned.
In such a way the function returns $+1$ in a 'small' region (capturing the training data points) and $-1$ elsewhere.

\subsubsection{Data preprocessing}

\begin{table}
  \centering
  \caption{Properties of training and testing sets used with \gls{OC-SVM} with respect to the \var{window\_size}.}
  \label{tab:experiments:svm_data}
  \begin{tabular}{llccc}
    \toprule
    & & \multicolumn{3}{c}{\var{window\_size}} \\
    \cmidrule{3-5}
    & & \num{1024} & \num{512} & \num{128} \\
    \midrule
    \multicolumn{2}{l}{Training set samples} & \num{74607} & \num{149217} & \num{596875}\\ 
    \midrule
    \multirow{3}{*}{\rotatebox[origin=c]{90}{\parbox[c]{5.2em}{\centering Testing set (augmented)}}} & \makecell[l]{initial no of\\examples} & \num{9350} & \num{18702} & \num{74812} \\
    & \makecell[l]{quenches\\multiplication} & \num{4675} & \num{9351} & \num{37406} \\
    & \makecell[l]{final no of\\examples} & \num{14025} & \num{28053} &  \num{112218} \\
    \bottomrule
  \end{tabular}
\end{table}


The \gls{OC-SVM} was trained on the normal operation data.
As described in subsection \ref{subsection:experiments:dataset}, four series (h1011, h1144, h1451 and h1819) were used, all coming from the same magnet, with only the actual voltage values for each of the signals used in experiments.
Each of the series was then split into two parts: a training set, containing only normal operation data and a testing set containing the anomalies.

The sets were then preprocessed and various features extracted, similar to \citep{s141120713}, to achieve a simpler classifier.
The extracted features and their properties are represented in \ref{appx:parameters} in Tab.~\ref{tab:tab_feature_desc}.
Tab.~\ref{tab:experiments:svm_data} shows different \var{window\_size} values (in samples) which were used for extracting the features and the resulting training and testing sets.

\subsubsection{Training and testing}

\begin{table}
  \centering
  \caption{\gls{OC-SVM} performance results.}
  \label{tab:experiments:detector_performance_svm}
  \begin{tabular}{llccc}
    \toprule
    & & \multicolumn{3}{c}{\var{window\_size}} \\
    \cmidrule{3-5}
    & & \num{1024} & \num{512} & \num{128} \\
    \midrule
  	\multirow{5}{*}{Training set} & accuracy & \num{0.94} & \num{0.98} & \textbf{\num{0.99}} \\
    & recall & \num{1.0} & \num{1.0} & \num{1.0} \\
    & precision & \num{1.0} & \num{1.0} & \num{1.0} \\
    & $\mathrm{F}_1$ & \num{1.0} & \num{1.0} & \num{1.0} \\
    & $\mathrm{F}_2$ & \num{1.0} & \num{1.0} & \num{1.0} \\ 
    \midrule
  	\multirow{5}{5.2em}{Testing set (augmented)} & accuracy & \textbf{\num{0.92}} & \textbf{\num{0.92}} & \num{0.87} \\
    & recall & \num{1.0} & \num{1.0} & \num{1.0} \\
    & precision & \num{1.0} & \num{1.0} & \num{1.0} \\
    & $\mathrm{F}_1$ & \num{1.0} & \num{1.0} & \num{1.0} \\
    & $\mathrm{F}_2$ & \num{1.0} & \num{1.0} & \num{1.0} \\ 
    \midrule
    \multirow{5}{5em}{Synthetic set~I} & accuracy & \num{0.72} & \num{0.85} & \textbf{\num{0.96}} \\
    & recall & \textbf{\num{1.0}} & \textbf{\num{1.0}} & \num{0.84} \\
    & precision & \textbf{\num{0.28}} & \num{0.15} & \num{0.03} \\
    & $\mathrm{F}_1$ & \textbf{\num{0.43}} & \num{0.25} & \num{0.06} \\
    & $\mathrm{F}_2$ & \textbf{\num{0.66}} & \num{0.46} & \num{0.13} \\ 
    \midrule
 	\multirow{5}{5em}{Synthetic set~II} & accuracy & \num{0.74} & \num{0.87} & \textbf{\num{0.95}} \\ 
    & recall & \textbf{\num{1.0}} & \textbf{\num{1.0}} & \num{0.88} \\
    & precision & \textbf{\num{0.26}} & \num{0.13} & \num{0.04} \\
    & $\mathrm{F}_1$ & \textbf{\num{0.42}} & \num{0.23} & \num{0.08} \\
    & $\mathrm{F}_2$ & \textbf{\num{0.64}} & \num{0.43} & \num{0.19} \\ 
    \bottomrule
  \end{tabular}
\end{table}

\gls{OC-SVM}, using \gls{RBF} kernel, was trained on the preprocessed training dataset (the grid search method to find the best values of $\nu$ ($=0.07$) and $\gamma$ ($=0.06$) was used).
After training, the model has run on the test datasets where there were data labeled with known quenches.
Due to very few numbers of quenches in the test data, the test set has been augmented, increasing the number of quench instances so that it would match normal samples cardinality.
Tab.~\ref{tab:experiments:svm_data} shows the properties of this augmented test set.
The model has also run on the same synthetic data sets that were used to validate the \gls{GRU}-based detector.
Fig.~\ref{fig:experiments:oc_svm_diagram} summarizes the main processing stages for \gls{OC-SVM} and contains a high-level description of the methods used in this alternative approach.

Tab.~\ref{tab:experiments:detector_performance_svm} shows the results (following the metrics described in \ref{subsection:experiments:quality-measures}) of the \gls{OC-SVM} on the \gls{HL} data with synthetic data sets.

\section{Discussion}
\label{section:discussion}

To be able to measure the detector system performance, it is necessary to answer the question of what and where an anomaly is in this context.
Exact anomaly position is tough to determine when it comes to the signals acquired from the \gls{LHC} magnets, especially the new \gls{HL} ones.

The target model should be trained using data acquired with the very high sampling rate, obtained during the experiments with the magnets.
This kind of data is not available during normal operation in any database because of limited network throughput. 
Therefore the target system can be applied only in the immediate vicinity of the signal source, directly on the detecting device located near superconducting component in the \gls{LHC} tunnel, as it is only place where data is available without any decimation.
Hardware implementation (in \gls{FPGA} or \gls{ASIC}) is also required to meet the \gls{CERN} requirement of very low system response latency.

The solution based on \glspl{NN} was selected because those models may be updated automatically, require minimal feature extraction, can be compressed efficiently and ported into hardware \citep{chang2015recurrent, han2017ese, lee2016fpga}.

Choosing the right model for the task and its hyper-parameters adjusting can be a very time- and resource-consuming process.
An automation of that process should potentially not only adjust the model hyper-parameters but also address the problem of model compression/precision reduction challenges and therefore make a \gls{NN}-based solution hardware implementation much more manageable.
The currently used at \gls{CERN} anomaly detection system required the adoption of the high-level models using \gls{HDL} (e.g., VHDL).
Such an approach results in a complicated and error-prone process.
Furthermore, any updates or modifications of the high-level models require a complete reiteration of the design flow. 
The adaptability of the \gls{NN}-based system coupled with the automatic optimization algorithm could significantly simplify that process.

It is also worth noting that historically at \gls{CERN} feature extraction was a challenging phase since it involved many experiments with a range of filtering and discrimination methods to reach reliable parameters of the system as a whole.
While the naive adoption of \glspl{RNN} requires an operator of the system to make an arbitrary decision regarding the values of the thresholds \citep{wielgosz2017lstm}, the adaptive inputs and outputs quantization approach presented in this work alleviates this issue by introducing an automatic required analyzer thresholds adjustment process.

\section{Conclusions}
\label{section:conclusions}

In this paper, an applicability of \gls{RNN} models for detecting anomalous behavior of \gls{CERN} superconducting magnets was examined.
The developed solution, based on \gls{GRU} and adaptive quantization, achieved very encouraging results for the data acquired from \gls{HL} magnets.
Three testing sets were used in the experiments, one including real anomalies and two with synthetic anomalies in the form of a unit step impulse with the length of 100 and 50 samples.
For those datasets, the proposed anomaly detection system reached $\mathrm{F}_2$ equal to \num{1.0}, \num{0.9980}, and \num{0.8920}, respectively, with $\mathrm{F}_1$ equal to \num{1.0}, \num{0.9987}, and \num{0.9296}.
Several setups of the proposed solution were analyzed, with the configurations selected based on shortest reported false anomaly length and best underlying model accuracy achieving the best results.

An essential part of the proposed solution is the adaptive quantization algorithm.
It can convert 20-bits input samples to reduced (e.g., 4-bits) representation that can be used as the model input.
The input and output quantization parameters turned out to have a significant impact on the detector performance, with 16/8 ratio providing the best overall results among the tested cases.
Another noteworthy aspect of the developed anomaly detector is specially designed analyzer which processes anomalies' candidates.  

Despite being primarily focused on the \gls{CERN} equipment monitoring, the results presented in this paper should be considered a part of a larger endeavor aiming at developing a methodology and architecture of an anomaly detection system operating in the space of time series analysis, especially under hard real-time constraints.

\section{Future work}
\label{section:future}
As a future work, the authors plan to further test and improve the proposed algorithm.
It is planned to examine the proposed solution performance using more advanced quality measures and more sophisticated testing datasets and improve adaptive quantization, automatic analyzer rules selection and anomaly discrimination algorithms.

Data with even higher sampling rate will be used.
It is worth noting that this data contains even more significant noise component produced by power converters delivering current to coils.
It can also contain other fast physical phenomena (flux jumping) that can take place inside a superconductor.
Correct anomaly detection must be performed despite these phenomena.
Further research can be done regarding a possibility of early warnings (before anomaly) and the existence of quench precursors.

Ultimately, the authors plan to develop an \gls{RL}-based \gls{NN} model optimization algorithm, the preliminary idea of which was presented in \citep{wielgosz2016observer}, and use it to simplify the process of the detector prototype implementation on an \gls{FPGA} platform.

\bibliographystyle{elsarticle-num}
\bibliography{bibliography-cern,bibliography-anomaly-detection,bibliography-machine-learning,bibliography-misc,wielgosz-published,wielgosz-unpublished}

\appendix
\section{Software symbols}
\label{appx:soft_symb}

\begin{table*}
  \centering
  \caption{Software symbols used in the text.}
\label{tab:tab_soft_symb}
  \begin{tabular}{m{5.6em}>{\raggedright}p{9.4em}p{28em}}
    \toprule
Section & Variable name & Description \\
  \toprule
\multirow{12}{5.6em}{data preprocessing} & \var{look\_back}	& the history window length \\ 
& \var{look\_ahead} & the gap between last historical sample and predicted one; in presented experiments \var{look\_back}{=1} (predict sample immediately following the history) \\ 
& \var{in\_grid} & the number of input data quantization levels \\ 
& \var{out\_grid} & the number of output data quantization levels \\ 
& \var{in\_algorithm} & the input data quantization algorithm, can be \textit{static}, \text{adaptive} (used in presented experiments) or  \textit{none} (no quantization) \\ 
& \var{out\_algorithm} & analogical to \var{in\_algorithm}, but for output data \\ 
& \var{samples\_percentage}	 & the percentage of total available data the model is trained on; it is determined during preprocessing and all following \obj{model} and \obj{analyzer} instances are using the same subset \\ 
    \midrule
\obj{model} & \var{cells} & the number of \gls{GRU} model cells; presented experiments utilized only single \gls{GRU} layer, but it is possible to specify more \\ 
    \midrule
\multirow{7}{*}{\obj{analyzer}} & length threshold$^\dagger$ & an anomaly candidate length (in samples) that qualifies it as an anomaly  \\
 & maximum amplitude threshold$^\dagger$ & an anomaly candidate maximum amplitude (measured as a distance between real and predicted sample quantization bin middles) that qualifies it as an anomaly \\ 
 & cumulative amplitude threshold$^\dagger$ & sum of anomaly candidate amplitudes that qualifies it as an anomaly \\ 
    \bottomrule
  \end{tabular}
\flushleft
\footnotesize{$^\dagger$various threshold values can be combined, creating a set of rules allowing to determine if an anomaly candidate is an anomaly}
\end{table*}

The Tab.~\ref{tab:tab_soft_symb} summarizes the names used in the developed software \citep{online:bitbucket:anomaly_detection} and useful for understanding this text.

\section{Statistical parameters}
\label{appx:parameters}

\begin{table*}
  \centering
  \caption{Feature extractors used for \gls{OC-SVM} \citep{s141120713, 4463657}. The input signal is denoted as $x[n]$, where $n=\{1,2,\dots,N=\var{window\_size}\}$.}
  \label{tab:tab_feature_desc}
  \begin{tabular}{>{\raggedright}m{8em}m{18em}m{17em}}
    \toprule
    Feature & Equation & Description \\ 
    \midrule
    Average Power & $\overline{P} = \frac{1}{N}\sum_{n=1}^{N}x[n]^{2}$ & overall vibration intensity of the window \\
    Mean Value & $\overline{x}=\sum_{n=1}^{N}x[n]$ & amplitude of low frequency \\
    Median frequency & $MF = \begin{cases} (\frac{n+1}{2}) &\mbox{if } n \equiv odd \\
    \frac{(\frac{n}{2}) + (\frac{n+1}{2})}{2} & \mbox{if } n \equiv even \end{cases}$ & frequency of the power spectrum into two halves with the same energy \citep{4463657} \\
    Standard deviation & $\sigma = \sqrt{\frac{1}{N-1}\sum_{n=1}^{N}(x[n]-\overline{x})^{2}}$ & shape of the signal \\
    Skewness & $s_{o}=\frac{\sqrt{N(N-1)}}{N-2}\frac{\frac{1}{N}\sum_{n=1}^{N}(x[n]-\overline{x})^{3}}{\left(\sqrt{({1}{N}\sum_{n=1}^{N})(x[n]-\overline{x})^{2}}\right)^{3}}$ & reflects asymmetries \\
    Kurtosis & $k_{o}=\frac{N-1}{(N-2)(N-3)}\left((N+1)k_{1}-3(N-1)\right)+3$, where $k_{1}=\frac{\frac{1}{N}\sum_{n=1}^{N}(x[n]-\overline{x})^{4}}{\left(\frac{1}{N}\sum_{n=1}^{N}x[n]-\overline{x})^{2}\right)^{2}}$ & peakedness of the histogram \\
    Central Tendency Measurement & CTM $=$ first-order differences on scatter plot representing $x[n+1]-x[n]$ on the X axis against $x[n+2]-x[n+1]$ on the Y axis & randomness of the signal, low value implies sharp changes \\
    Correlation coefficient & $r = \frac{\sum_{n=1}^{N-2}(X[n]-\overline{X})(Y[n]-\overline{Y})])}{\sqrt{\sum_{n=1}^{N-2}(X[n]-\overline{X})^{2}} \sqrt{\sum_{n=1}^{N-2}(Y[n]-\overline{Y})^{2}}}$ & unpredictability of the signal   from previous data; correlation between the first-order differences of scatter plot with the Pearson's linear correlation \\
    Lempel-Ziv Complexity & $LZC = \frac{L(N)}{N}$ where $L(N) \equiv $ length of the encoded sequence & characterizes the average information quantity within a window \citep{4463657} \\
    \bottomrule
  \end{tabular}
\end{table*}

The Tab.~\ref{tab:tab_feature_desc} list well-known statistical parameters. In this study they were applied for feature extraction in reference approach based on \gls{OC-SVM} method.

\section{GRU}
\label{appx:gru}

\begin{figure}
  \centering
  \includegraphics[width=0.75\hsize]{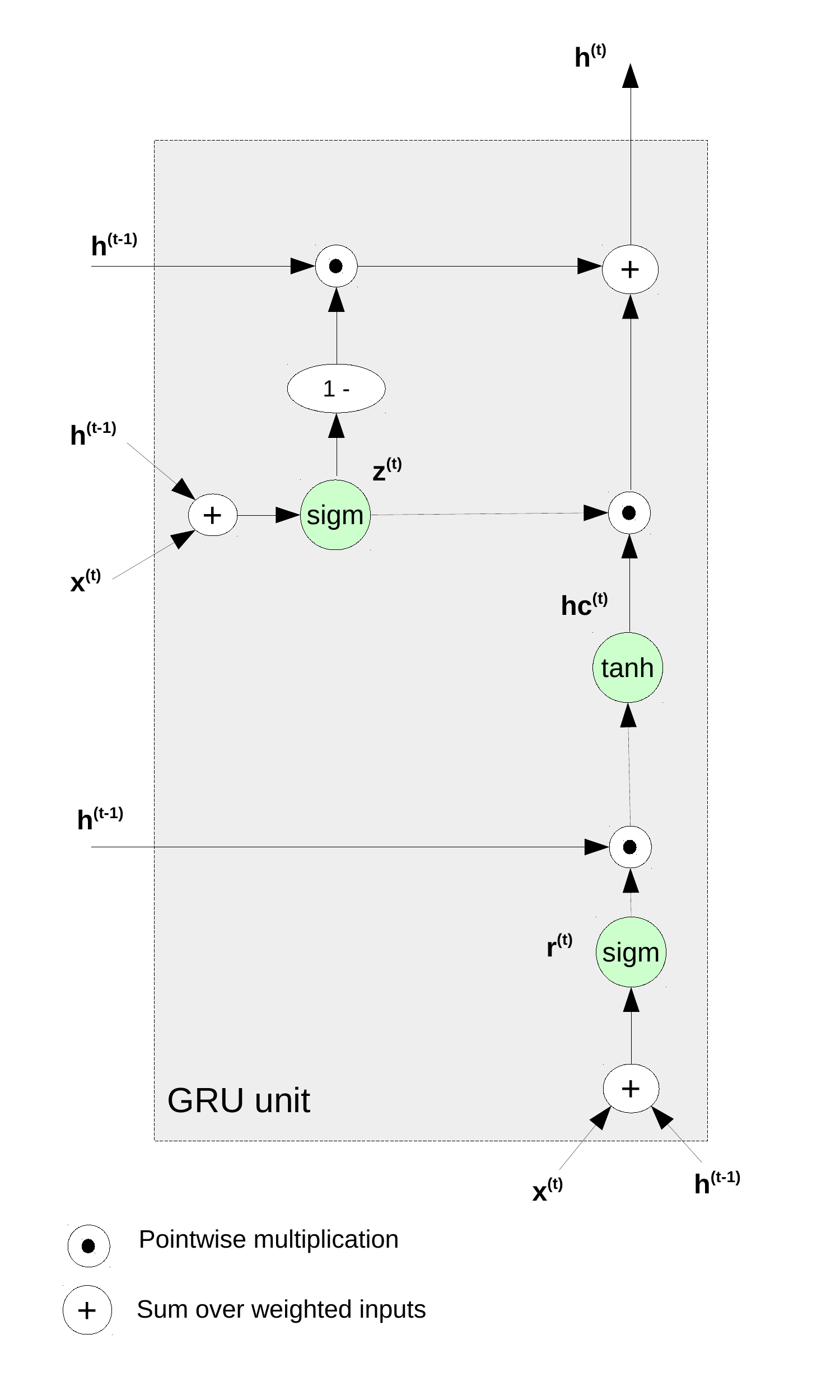}
  \caption{Architecture of \gls{GRU} unit.}
  \label{fig:rnns:gru_unit_architecture}
\end{figure}

The \gls{GRU} has gating components which modulate the flow of information within the unit, as presented in Fig.~\ref{fig:rnns:gru_unit_architecture}.
In the related equations (\ref{eq:gru_h}) to (\ref{eq:gru_hc}) a sigmoidal function is denoted by symbol $\sigma$, and an hyporbelic tangent is denoted as $\phi$.

\begin{equation}
h^{(t)} = (1 - z^{(t)}) \odot h^{(t-1)} + z^{(t)} \odot hc^{(t)}.
 \label{eq:gru_h}
\end{equation}

The activation of the model at a given time $t$ is a linear interpolation between the activation $h^{(t-1)}$ from the previous time step and the candidate activation $hc^{(t)}$.
It is desribed by equation (\ref{eq:gru_h}) above. The activation is strongly modulated by quantity $z^{(t)}$ as given by (\ref{eq:gru_h_unfolded_0}) and  (\ref{eq:gru_h_unfolded_1}):

\begin{equation}
\begin{split}
h^{(t)} & = (1 - z^{(t)}) \odot h^{(t-1)} + z^{(t)} \odot hc^{(t)} \\
& = h^{(t-1)} - z^{(t)} \odot h^{(t-1)} + z^{(t)} \odot hc^{(t)},
\end{split}
 \label{eq:gru_h_unfolded_0}
\end{equation}

\begin{equation}
h^{(t)} =  h^{(t-1)} - z^{(t)} \odot (h^{(t-1)} + hc^{(t)}).
 \label{eq:gru_h_unfolded_1}
\end{equation}

\begin{equation}
z^{(t)} = \sigma(W_{zx}x^{(t)} + W_{zh}h^{(t-1)}).
 \label{eq:gru_z}
\end{equation}

The formula for the update gate is given by (\ref{eq:gru_z}) and modulates a degree to which a \gls{GRU} unit updates its activation. 
The \gls{GRU} has no mechanism to control to what extent its state is exposed, but it exposes the whole state each time. 

\begin{equation}
r^{(t)} = \sigma(W_{rx}x^{(t)} + W_{rh}h^{(t-1)}).
 \label{eq:gru_r}
\end{equation}

The response of the reset gate is computed according to the same principle as the update gate.
The previous state information $h^{(t-1)}$ is multiplied by the coefficients matrix $W_{rh}$ and the input data $x^{(t)}$ is multiplied by the coefficients matrix $W_{rx}$ as shown in (\ref{eq:gru_r}).

\begin{equation}
hc^{(t)} = \phi(W_{hh}r^{(t)} \odot h^{(t-1)} + W_{hx}x^{(t)}).
 \label{eq:gru_hc}
\end{equation}

The candidate activation $hc^{(t)}$ is computed according to  (\ref{eq:gru_hc}).
When $r^{(t)}$ is close to 0 (meaning that the gate is almost off), the stored state is forgotten.
The input data $x^{(t)}$ is read instead.

As it was pointed out, \gls{GRU} has a simpler structure than \gls{LSTM} \cite{wielgosz2017lstm, wielgosz2017recurrent} which is also reflected in the performance of the algorithm.

\end{document}